%% file: iclr2027_conference.tex
\documentclass{article}
\usepackage{iclr2027_conference,times}

\input{math_commands.tex}

\usepackage{url}
\usepackage{float}
\usepackage{wrapfig}
\usepackage{booktabs}
\usepackage{hyperref}
\usepackage{multirow}
\usepackage[table]{xcolor}
\usepackage{graphicx}
\usepackage{comment}
\definecolor{rowmint}{RGB}{232,243,241}

\title{MARS: Systematic Multi-Level Acceleration for World Action Models}

\author{\textbf{Zengyuan Zuo}$^{1,2*}$\quad
\textbf{Yongchang Zhang}$^{1,2*}$\quad
\textbf{Guangyu Zhao}$^{1,2}$\quad
\textbf{Tianze Zhang}$^{1,2}$\\
\textbf{Biao Wu}$^{3}$\quad
\textbf{Meng Fang}$^{4}$\quad
\textbf{Yiwu Zhong$^{1,2\dag}$}\\
\textsuperscript{\rm 1}School of Intelligence Science and Technology, Peking University\\
\textsuperscript{\rm 2}State Key Laboratory of General Artificial Intelligence, Peking University\\
\textsuperscript{\rm 3}University of Technology Sydney\quad
\textsuperscript{\rm 4}University of Liverpool\\
}

\iclrfinalcopy
\begin{document}

{
  \renewcommand{\thefootnote}{\fnsymbol{footnote}}
  \footnotetext[1]{Equal contribution. \textsuperscript{\dag}Corresponding author.}
}

\maketitle

\input{sections/abstract}

\input{sections/sec_introduction_zyc}

\input{sections/sec_related_work}

\input{sections/method_final_v2}

\input{sections/sec_experiments}

\input{sections/sec_conclusion}

\bibliography{iclr2027_conference}
\bibliographystyle{iclr2027_conference}
\input{sections/appendix}

\end{document}

%% file: math_commands.tex
\usepackage{amsmath,amsfonts,bm}

\def\eqref#1{equation~\ref{#1}}
\def\1{\bm{1}}

\DeclareMathAlphabet{\mathsfit}{\encodingdefault}{\sfdefault}{m}{sl}
\SetMathAlphabet{\mathsfit}{bold}{\encodingdefault}{\sfdefault}{bx}{n}

%% file: sections/abstract.tex
\begin{abstract}
World Action Models (WAMs) have emerged as a promising alternative for embodied intelligence by jointly predicting future world states and robot actions. However, their heavy visual encoding and iterative generative inference incur substantial latency, limiting their applicability to high-frequency closed-loop control. Existing training-free acceleration methods typically optimize a single stage, yielding limited speedups or degraded control under aggressive reuse.
In this work, we systematically profile the inference overhead of WAMs and observe abundant redundancy across Transformer computation, diffusion denoising, and consecutive policy queries. However, the safe removal of such redundancy depends strongly on its impact on action prediction and control sensitivity.
Based on these observations, we introduce \textbf{MARS}, a training-free and plug-and-play framework for multi-level WAM acceleration. MARS selectively removes low-impact attention computation, adaptively accelerates diffusion inference according to motion granularity, and speculatively reuses predicted future visual representations across consecutive queries with online verification.
Together, these complementary mechanisms provide remarkable acceleration while limiting approximation error. Extensive experiments across LIBERO, LIBERO-PRO, and RoboCasa show that MARS achieves up to $4.00\times$, $3.94\times$, and $2.10\times$ end-to-end speedups, respectively, while maintaining task success rate close to the original policy. 
Moreover, MARS generalizes across different model backbones and across simulation and real-robot experiments, substantially improving the efficiency--performance trade-off of WAM inference. Our code is available at \url{https://github.com/PKU-VaLuE-Lab/MARS}.
\end{abstract}

%% file: sections/sec_introduction_zyc.tex
\section{Introduction}
\label{sec:introduction}

World Action Models (WAMs) have recently emerged as a promising foundation-model paradigm for embodied intelligence~\citep{guo2024prediction,zhu2025unified,shen2026videovla,kim2025cosmospolicy,ye2026world}. Unlike Vision-Language-Action (VLA) models that directly map observations to robot actions~\citep{kim2024openvla,intelligence2025pi_}, WAMs jointly generate future world states and actions within a unified generative process. By explicitly modeling how the environment evolves under robot interaction, WAMs can leverage physical priors and interaction dynamics acquired from large-scale video pretraining~\citep{wu2024unleashing,agarwal2025cosmos}, offering a compelling path toward more generalizable robotic policies.

This additional modeling capability, however, comes at a substantial computational cost. A WAM policy query typically requires both encoding high-dimensional visual observations and performing iterative generation through a large diffusion Transformer. As a result, WAM inference can be considerably slower than conventional action models. In practice, this latency often forces robots to execute longer action chunks and replan less frequently, weakening closed-loop responsiveness and limiting the practical benefits of WAMs for high-frequency control. Efficient inference is therefore essential for translating their modeling advantages into effective robotic behavior.

A natural approach is to directly adapt existing training-free acceleration techniques developed for diffusion models or robotic policies~\citep{ma2024deepcache,bu2025dicache,zhao2026c,zeng2026x}. Yet such methods typically exploit redundancy along only a single computational dimension, inevitably facing an efficiency-performance trade-off ceiling. 
Conservative diffusion caching can preserve policy quality, but usually yields only moderate latency reduction. More aggressive reuse across consecutive action chunks can remove substantially more computation, but approximation errors may accumulate over time and severely degrade long-horizon task success. 
More importantly, WAM inference contains multiple expensive stages whose relative costs change after optimization: once the dominant DiT computation is substantially reduced, visual encoding can immediately become the new latency bottleneck. Consequently, isolated, single-stage optimizations are insufficient for sustained end-to-end acceleration, calling for a holistic solution.

\begin{figure}[t]
\centering
\includegraphics[width=\textwidth]{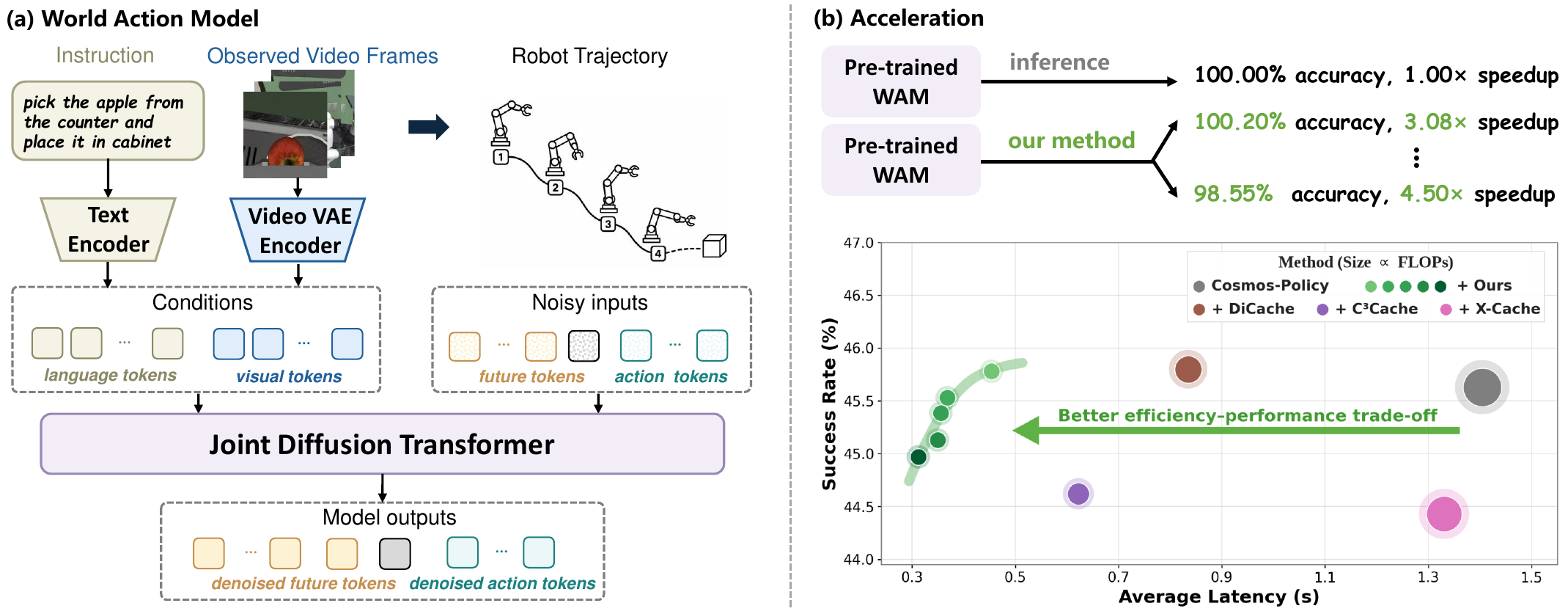}
\vspace{-4mm}
\caption{\textbf{WAM inference and acceleration overview.} (a) The overall pipeline of a World Action Model. (b) Efficiency-performance comparison in LIBERO-PRO~\citep{zhou2025libero}, where our method substantially reduces latency and FLOPs with a better efficiency–performance trade-off.}
\label{fig:overview}
\vspace{-4mm}
\end{figure}

These limitations raise two central questions: \emph{where does redundant computation arise in WAM inference, and which computations can be safely removed?} To answer them, we systematically profile the WAM inference pipeline, as illustrated in Figure~\ref{fig:overview}(a), and obtain three observations.
\textbf{First}, redundancy within the Transformer is highly non-uniform. The contributions of individual attention components vary substantially: many cross-attention branches and self-attention heads have little influence on the final action prediction, whereas a relatively small subset remains critical. This suggests that attention computation should be reduced selectively rather than uniformly.
\textbf{Second}, redundancy across diffusion steps depends on the control stage. Although the sampler input changes substantially across denoising steps, the output states and  DiT residuals can remain comparatively stable, indicating considerable opportunity for computation reuse. However, different manipulation stages exhibit markedly varying sensitivities to approximation: coarse motions are often tolerant to aggressive acceleration, while fine-grained interactions such as grasping and placement require more accurate iterative refinement.
\textbf{Third}, redundancy also exists across consecutive policy queries, but naive reuse is unreliable. Since a WAM explicitly predicts future visual states, the predicted visual representation from the current query can potentially serve as the visual input for the next query, avoiding repeated encoding of nearly identical observations. However, prediction errors and unexpected environment changes make unconditional reuse unsafe, especially over long horizons.

Together, these observations reveal a distinctive characteristic of WAM inference: computational redundancy is abundant, but it is distributed non-uniformly across network components, denoising steps, and consecutive queries, and removing it at different locations incurs varying risks to action control. 
Effective acceleration should therefore not rely on a single reuse rule or optimize one stage in isolation. Instead, computation should be reduced according to its importance to action prediction, while aggressive reuse that can introduce temporally accumulated errors should be validated online. In short, the key is not merely to remove more computation, but to remove the \emph{right} computation at the \emph{right} stage with respect to the control sensitivity.

Guided by this principle, we propose \textbf{MARS}, a training-free and plug-and-play framework that accelerates WAM inference at three complementary levels. 
At the \textbf{Transformer level}, MARS uses calibration statistics to skip low-impact cross-attention branches and prune low-importance self-attention heads while retaining components critical to action prediction. 
At the \textbf{diffusion level}, MARS uses the provisional action produced after the first denoising step to estimate the motion granularity type. For coarse-grained motions that tolerate approximation, it performs one-step generation. For fine-grained stages, it preserves the original multi-step solver while reusing stable DiT residuals to reduce backbone computation. 
Finally, at the \textbf{cross-query level}, MARS treats the predicted future visual tokens as a speculative representation of the next observation. It compares predicted and observed visual changes using shallow VAE features and reuses the predicted tokens only when they are sufficiently consistent, thereby avoiding unnecessary visual encoding.

We evaluate MARS extensively across multiple robotic manipulation benchmarks. While maintaining task success rate close to that of the original policy, MARS achieves up to $4.00\times$ end-to-end speedup on LIBERO~\citep{liu2023libero}, $3.94\times$ on LIBERO-PRO~\citep{zhou2025libero}, and $2.10\times$ on RoboCasa~\citep{nasiriany2024robocasa}, together with substantial reductions in FLOPs. 
By independently adjusting the acceleration strength at the three levels, MARS further enables flexible efficiency--performance trade-offs, as illustrated in Figure~\ref{fig:overview}(b). 
Moreover, its applicable components consistently accelerate different WAM and VLA backbones, suggesting that the identified forms of redundancy are not specific to a particular architecture.

Our contributions are summarized as follows:

\textbullet\ \ We systematically profile WAM inference and reveal three properties: redundancy spans multiple stages, safe removal is stage-dependent, and acceleration shifts the computational bottleneck.

\textbullet\ \ We introduce MARS, a training-free, plug-and-play framework that jointly exploits redundancy in Transformer computation, diffusion denoising, and cross-query visual processing.

\textbullet\ \ Extensive experiments across benchmarks, base models, hardware platforms and real-world tasks demonstrate that MARS achieves a substantially improved efficiency–performance trade-off.

%% file: sections/sec_related_work.tex
\section{Related Work}
\label{sec:related}

\textbf{World Action Models.} World Action Models (WAMs), an emerging class of embodied foundation models, explicitly model how the environment evolves under candidate actions. They either jointly model future states and actions~\citep{guo2024prediction,zhu2025unified,shen2026videovla,kim2025cosmospolicy,ye2026gigaworld} or predict future states and derive actions from them~\citep{du2023learning,pai2025mimic,feng2025vidar}. From a generative modeling perspective, existing WAMs are primarily instantiated using autoregressive~\citep{wu2024unleashing,zhao2025cot,cen2025rynnvla} or diffusion-based formulations. We focus on diffusion-based WAMs. Unified World Models~\citep{zhu2025unified} jointly model videos and actions through modality-aware diffusion. Cosmos Policy~\citep{kim2025cosmospolicy} adapts a pretrained video diffusion backbone~\citep{agarwal2025cosmos} to jointly generate future images, actions, proprioceptive states, and values. DreamZero~\citep{ye2026world} demonstrates that large video-pretrained WAMs can serve as zero-shot policies with strong transfer to unseen motions and embodiments. These works demonstrate the effectiveness of future prediction for action generation, but their video-generation backbones introduce substantial deployment costs.

\textbf{Efficient World Action Models.} Recently, several works have explored reducing the inference cost of WAMs. Fast-WAM~\citep{yuan2026fast} and GigaWorld-Policy~\citep{ye2026gigaworld} retain video co-training but remove explicit future generation at test time. Efficient-WAM~\citep{li2026efficient} reduces future-imagination cost through a compact video expert, low-resolution future latents, and asymmetric video-action denoising. Flash-WAM~\citep{akbari2026flash} instead adopts modality-aware consistency distillation to compress both video and action denoising to a single step. Despite achieving substantial speedups, these approaches require dedicated training or architectural changes. Among training-free approaches, C$^3$ache~\citep{zhao2026c} and X-Cache~\citep{zeng2026x} exploit cross-chunk similarity to reuse intermediate residuals, with X-Cache further employing structure- and action-aware fingerprints for block-level cache matching. Dream-Tac~\citep{lou2026dream} instead applies a fixed cache schedule to selected denoising steps within each action chunk. These methods primarily exploit redundancy at a single computational level, whereas we jointly address multi-level redundancy in attention computation, diffusion denoising, and observation encoding.

\textbf{Training-Free Acceleration for Diffusion and VLA Models.}
Training-free diffusion acceleration commonly reduces redundancy across denoising steps and network components. DeepCache~\citep{ma2024deepcache} reuses intermediate features with predefined schedules, while TeaCache~\citep{liu2025timestep} and DiCache~\citep{bu2025dicache} estimate feature changes online. At a finer granularity, ToCa~\citep{zou2024accelerating}, PAB~\citep{zhao2025real}, and DiTFastAttn~\citep{yuan2024ditfastattn} reduce token- or attention-level computation. WorldCache~\citep{nawaz2026worldcache} further introduces motion- and saliency-aware reuse for video world models. However, these methods are primarily optimized for perceptual fidelity rather than control accuracy. For robot policies, BAC~\citep{ji2025block} exploits block-wise redundancy, while DP-Cache~\citep{zhou2026characterizing}, V-AEFusion~\citep{zhou2026characterizing}, and ActionCache~\citep{oi2026training} accelerate action generation through caching, computation overlap, or warm-starting. VLA-Cache~\citep{xu2026vla} and EfficientVLA~\citep{yang2026efficientvla} reduce visual computation through temporal reuse or token selection, whereas SP-VLA~\citep{li2026spvla}, SpecPrune-VLA~\citep{wang2025specprune}, and ADP~\citep{pei2025action} allocate computation according to action dynamics or manipulation stages. Nevertheless, these VLA acceleration methods largely focus on the VLM backbone, leaving the multi-level redundancy in WAM inference underexplored.

%% file: sections/method_final_v2.tex
\section{Method}
\label{sec:method}

\subsection{Preliminaries \& Problem Formulation}

WAMs jointly generate future world states and actions within a unified generative process. We use Cosmos Policy as the base WAM, and Figure~\ref{fig:overview}(a) shows its pipeline. At each policy query $q$, Cosmos Policy takes multi-view observations $\mathcal{I}_q=\{I_q^c\}_{c=1}^{C}$, robot proprioception $p_q$, and an instruction $\ell$ as input. It predicts
\begin{equation}
\left(A_q,\hat{\mathcal{Z}}_q^{\mathrm{future}},
\hat{p}_q^{\mathrm{future}},\hat{V}_q\right)
=
\mathcal{F}_{\theta}(\mathcal{I}_q,p_q,\ell),
\end{equation}
where $A_q$ is the action chunk, $\hat{\mathcal{Z}}_q^{\mathrm{future}}=\{\hat z_q^{c,\mathrm{future}}\}_{c=1}^{C}$ are the future visual tokens, $\hat{p}_q^{\mathrm{future}}$ is the future proprioception, and $\hat{V}_q$ is the value. For visual encoding, a VAE encoder $E$ encodes each view into visual tokens $z_q^c=E(I_q^c)$. A T5-XXL encoder~\citep{raffel2020exploring} encodes the instruction into language embeddings $e_\ell$, which are injected via cross-attention. A Diffusion Transformer (DiT) with $L$ Transformer blocks then generates all outputs through $K$ denoising steps. After the robot executes $A_q$, the next policy query starts with updated observations.

\subsection{Profiling and Diagnosing of WAM Redundancy}
\label{sec:insight}

\begin{figure}[H]
	\centering
	\includegraphics[width=\textwidth]{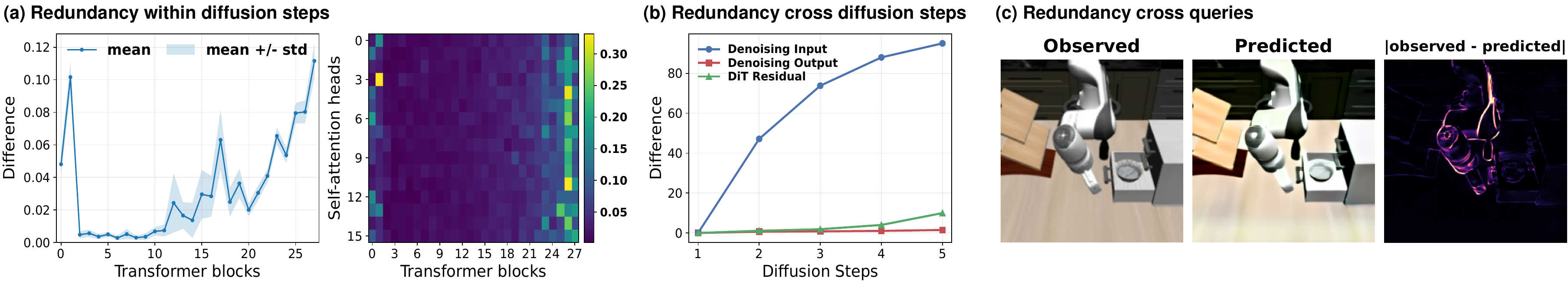}
	\caption{\textbf{Redundancy across multiple levels of WAM inference.} 
    \textbf{(a) Analysis 1}: Importance scores are measured by the effects after removing certain components. We find that cross-attention branches at early-to-middle layers exhibit low scores (e.g., blocks 2--11) and only few attention heads show high scores (e.g., yellow entries). 
    \textbf{(b) Analysis 2}: Relative L1 differences between denoising steps and the first denoising step. The denoising input varies substantially across steps, whereas the denoising output and DiT residual remain relatively stable. 
    \textbf{(c) Analysis 3}: Comparison of observed and predicted images. The prediction closely matches the observation with minor variations.}
	\label{fig:insight}
\end{figure}

We profile the WAM inference pipeline to locate redundant computation. We analyze three levels of model feedforward: the Transformer level (attention components within one DiT forward pass), the diffusion level (denoising steps within one policy query), and the cross-query level (visual encoding across consecutive policy queries). Figure~\ref{fig:insight} shows the analysis for each level. All diagnostics are performed with Cosmos Policy in the LIBERO~\citep{liu2023libero} simulation environment.

\textbf{Analysis 1: Transformer-Level Redundancy -- Non-Uniform Attention.}
We first investigate whether all attention components contribute equally within a DiT forward pass. We measure the importance of each component by the output change when the component is removed. 
For the cross-attention branch in block $b$, denote $x^{b,\mathrm{cross}}$ as its input and $y^{b,\mathrm{cross}}$ as its output. Removing the cross-attention branch leads to only the residual connection left and thus the output change equals to $y^{b,\mathrm{cross}}-x^{b,\mathrm{cross}}$. 
For self-attention head $h$ in block $b$, let $x^{b,\mathrm{self}}$ and $y^{b,\mathrm{self}}$ denote the block input and output, respectively. 
$y^{b,\mathrm{self},-h}$ denotes the block output when head $h$ is removed. The importance scores are computed as:
\begin{equation}
\label{eq:attention_scores}
\begin{aligned}
s_b^{\mathrm{cross}}
&=
\mathbb{E}
\left[
\frac{
\left\| y^{b,\mathrm{cross}} - x^{b,\mathrm{cross}} \right\|_1
}{
\left\| x^{b,\mathrm{cross}} \right\|_1 + \varepsilon
}
\right],
\qquad
s_{b,h}^{\mathrm{self}}
=
\mathbb{E}
\left[
\frac{
\left\| y^{b,\mathrm{self}} - y^{b,\mathrm{self},-h} \right\|_1
}{
\left\| x^{b,\mathrm{self}} \right\|_1 + \varepsilon
}
\right],
\end{aligned}
\end{equation}
% where $\mathbb{E}$ averages over policy queries and denoising steps, and $\varepsilon$ is a small constant for numerical stability.
% Figure~\ref{fig:insight}(a) shows that the importance scores are highly non-uniform. Many cross-attention branches exhibit low importance scores, particularly those in blocks 2--11, while branches in the early and late blocks contribute more significantly. Similarly, only a subset of self-attention heads exhibits high importance scores, as indicated by the brighter (yellow) regions in the heatmap, whereas many other heads have limited influence on the block output. These observations suggest that substantial attention computation can be skipped with limited perturbation to intermediate representations, motivating selective rather than uniform attention skipping.
where $\mathbb{E}$ averages over policy queries and denoising steps, and $\varepsilon$ is a small constant for numerical stability.
As shown in Figure~\ref{fig:insight}(a), the importance scores are highly non-uniform. 
Many cross-attention branches exhibit low importance scores, particularly those in blocks 2--11, while the branches in the early and late blocks contribute more significantly. 
Similarly, only a subset of self-attention heads exhibits high importance scores, as indicated by the brighter entries in the heatmap. 
These observations suggest that substantial attention computation can be skipped by removing certain attention components, motivating a selective skipping strategy.

\textbf{Analysis 2: Diffusion-Level Redundancy -- Stable Output States and Residuals.}
We next test how the variables change across the denoising steps. 
Within a denoising step, the denoising input is the noisy latent fed into the DiT, while the denoising output is the flow-matching velocity predicted by the DiT. 
The denoising input is first processed through patch embedding layer and then becomes the Transformer input. After Transformer feedforward, the Transformer output is then processed by a timestep-conditioned final projection layer to produce the denoising output. 
Here, we compare three variables: the denoising input, the denoising output, and the DiT residual $R_{k}$ which is defined as the input-output difference of the Transformer at denoising step $k$.
For each variable $a_{k}$, we compute the normalized L1 difference compared to the first denoising step, $\|a_{k}-a_{1}\|_1/\|a_{1}\|_1$, averaged over policy queries and reported as percentages.

As shown in Figure~\ref{fig:insight}(b), the denoising input is increased substantially across diffusion steps, whereas the denoising output and the DiT residual remain comparatively stable. These observations indicate considerable opportunity for computation reuse. 
The stable denoising output motivates one-step generation. The small variation of the DiT residual suggests that the residual computed at the first denoising step can be reused to approximate subsequent steps. This strategy preserves the original multi-step solver and only reduces the backbone computation. 
Overall, these analyses motivate two acceleration modes: one-step generation and the multi-step solver with residual reuse.

\textbf{Analysis 3: Cross-Query Redundancy -- Speculative Visual Representation.}
At policy query $q-1$, the WAM predicts future visual tokens $\hat z_{q-1}^{c,\mathrm{future}}$ for each view $c$, representing the expected scene at the next policy query $q$. These predictions can serve as speculative visual representations of subsequent observations, potentially avoiding redundant visual encoding of similar observations.
To assess their accuracy, we compare the observed image $I_q$ with the predicted image decoded from $\hat z_{q-1}^{c,\mathrm{future}}$. As shown in Figure~\ref{fig:insight}(c), the minor difference between the two images indicate that the predicted visual representation closely matches the actual observation.

\begin{figure}[t]
\vspace{-4mm}
\centering
\includegraphics[width=\textwidth]{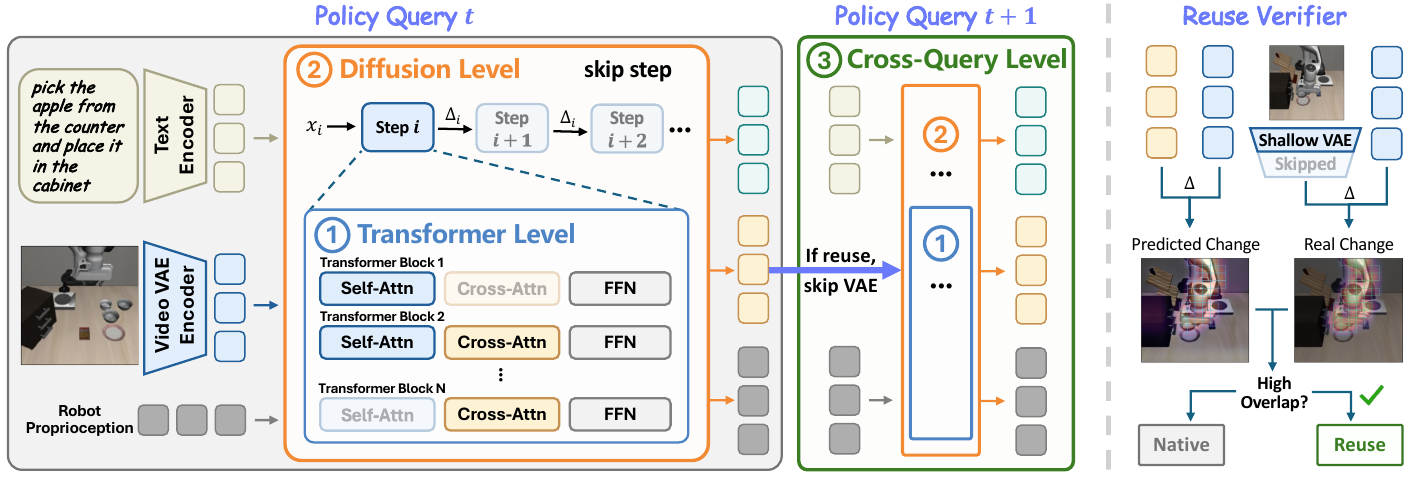}
\vspace{-6mm}
\caption{\textbf{The framework of MARS.} MARS is a training-free and plug-and-play framework that accelerates WAM inference at three complementary levels. At the Transformer level, it skips low-impact cross-attention branches and prunes low-importance self-attention heads. At the diffusion level, it uses the provisional action to estimate the motion granularity type and selects one-step generation or the multi-step solver with residual reuse. At the cross-query level, it reuses the predicted future visual tokens when they are sufficiently consistent with the observed visual changes.}
\label{fig:method}
\vspace{-3mm}
\end{figure}

\subsection{MARS Framework}

Motivated by these analyses, we propose MARS, a training-free and plug-and-play framework that accelerates WAM inference at three complementary levels, as shown in Figure~\ref{fig:method}. At the Transformer level, MARS reduces attention computation according to its importance to action prediction (Section~\ref{sec:method_attention}). At the diffusion level, it adapts the acceleration to the motion granularity type of each policy query (Section~\ref{sec:action_aware_stepskip}). At the cross-query level, it reuses the predicted future visual tokens when the online validation is passed (Section~\ref{sec:visual_token_reuse}).

\subsubsection{Transformer-Level Acceleration}
\label{sec:method_attention}

Based on Analysis 1, MARS removes the attention components with low importance scores. For cross-attention, it skips whole branches with low impacts. For self-attention, it prunes individual heads with low importance scores. 
% All components critical to action prediction are kept. 
We compute the importance scores in Eq.~\ref{eq:attention_scores} once offline and use them as calibration statistics. Given the numbers of skipped components $N_{\mathrm{cross}}$ and $N_{\mathrm{head}}$, the skip sets are
\begin{equation}
\label{eq:attention_skip_sets}
\mathcal{B}_{\mathrm{skip}}^{\mathrm{cross}}=\operatorname{BottomK}\left(\{s_b^{\mathrm{cross}}\}_{b=0}^{L-1},N_{\mathrm{cross}}\right),\quad
\mathcal{H}_{\mathrm{skip}}^{\mathrm{self}}=\operatorname{BottomK}\left(\{s_{b,h}^{\mathrm{self}}\}_{b,h},N_{\mathrm{head}}\right),
\end{equation}
where $\operatorname{BottomK}(\cdot,N)$ returns $N$ components with the lowest importance scores. For each block $b$ in $\mathcal{B}_{\mathrm{skip}}^{\mathrm{cross}}$, we skip the entire cross-attention residual branch:
\begin{equation}
\label{eq:cross_attention_skip}
\hat y^{b,\mathrm{cross}}
=
\begin{cases}
x^{b,\mathrm{cross}},
& b\in\mathcal{B}_{\mathrm{skip}}^{\mathrm{cross}},\\[1mm]
x^{b,\mathrm{cross}}
+
\operatorname{CA}_b
\left(
x^{b,\mathrm{cross}},e_\ell
\right),
& \text{otherwise},
\end{cases}
\end{equation}
where $\operatorname{CA}_b$ denotes the cross-attention layer in block $b$. For self-attention, we structurally prune the selected heads before attention aggregation:
\begin{equation}
\label{eq:self_attention_skip}
    \widetilde{\operatorname{MSA}}_b(X)
    =
    \sum_{
        h:\,(b,h)\notin
        \mathcal{H}_{\mathrm{skip}}^{\mathrm{self}}
    }
    \operatorname{head}_{b,h}(X)
    W_{O,b}^{(h)},
\end{equation}
where $X$ is the input to the self-attention layer in block $b$, $\operatorname{head}_{b,h}(X)$ is the output of head $h$, and $W_{O,b}^{(h)}$ is the output projection slice associated with head $h$.  % Removing head $h$ in Eq.~\ref{eq:attention_scores} follows the same operation.

\subsubsection{Diffusion-Level Acceleration}
\label{sec:action_aware_stepskip}

As shown in Analysis 2, the denoising output and the DiT residual keep almost the same after the first denoising step. 
However, how much approximation a policy query can tolerate depends strongly on the action motion. 
Coarse-grained motions are robust to approximation, while fine-grained interactions, such as grasping and placement, instead require precise iterative denoising. 
MARS therefore leverages the provisional action to estimate the motion granularity type of each policy query. 
For coarse-grained motions, it performs one-step generation. For fine-grained stages, it keeps the original multi-step solver, and the stable DiT residuals are reused to cut the backbone computation.

\textbf{Multi-Step Solver with Residual Reuse.}
Prior diffusion caching methods typically reuse intermediate features across consecutive denoising steps~\citep{ma2024deepcache,bu2025dicache}. To preserve perceptual quality in video generation, these methods introduce elaborate strategies to decide when to cache and how to reuse cached features. In WAMs, however, future visual prediction mainly serves as a dynamics prior for action generation. The ultimate objective is reliable control. We thus adopt a simpler caching strategy. We run the full DiT at the first denoising step and cache its residual $R_{q,1}$. At each remaining step, we skip the Transformer blocks and approximate their output by adding the cached residual to their current input. This mode preserves the original multi-step solver and only reduces the backbone computation after the first step. Following the original sampling schedule with these approximated outputs produces the residual-reuse action chunk $A_q^{\mathrm{RR}}$.

\textbf{Motion Granularity Estimation.}
The provisional action $A_q^{(1)}\in\mathbb{R}^{H\times7}$ is the action chunk produced after the first denoising step, where $H$ denotes the action-chunk horizon. It provides a natural signal for estimating the motion granularity type of the current policy query. We compute the mean magnitude of its translational components as
\begin{equation}
\label{eq:action_magnitude}
    m_q
    =
    \frac{1}{H}
    \sum_{i=1}^{H}
    \left\|
        A_{q,i,1:3}^{(1)}
    \right\|_2,
\end{equation}
where $A_{q,i,1:3}^{(1)}$ denotes the translational components of the $i$-th action. Large-motion chunks typically correspond to coarse-grained motions, such as reaching and transport. In contrast, small-motion chunks occur more frequently during fine-grained stages, such as approach, grasping, placement, and fine adjustment~\citep{wang2025specprune,li2026spvla}. We thus regard the motion as coarse-grained when $m_q > \tau_{\mathrm{act}}$ and as fine-grained otherwise. The final action chunk is
\begin{equation}
\label{eq:action_aware_step_skip}
    A_q
    =
    \begin{cases}
        A_q^{(1)}, & m_q > \tau_{\mathrm{act}}, \\[1mm]
        A_q^{\mathrm{RR}}, & m_q \leq \tau_{\mathrm{act}},
    \end{cases}
\end{equation}
where $\tau_{\mathrm{act}}$ controls the efficiency--performance trade-off. Unlike prior acceleration methods that infer manipulation stages from action history~\citep{wang2025specprune,li2026spvla}, MARS directly uses the provisional action of the current policy query. The acceleration thus adapts immediately to changes in motion granularity.

\subsubsection{Cross-Query Acceleration}
\label{sec:visual_token_reuse}

Once the Transformer-level and diffusion-level acceleration substantially reduce the dominant DiT computation, visual encoding becomes the new latency bottleneck. The proportion of total latency attributed to VAE encoding increases from 15.5\% to 47.7\% on LIBERO (Appendix~\ref{sec:latency}). Since a WAM explicitly predicts future visual tokens at each policy query, MARS treats the tokens predicted at query $q-1$ as a speculative representation of the next observation (Analysis 3). To validate this reuse online, our method first compares the predicted visual change with the observed visual change, which is measured using features from the shallow VAE layers. It reuses the predicted tokens only when the two changes are sufficiently consistent, thereby avoiding unnecessary visual encoding.

Specifically, let $E_{\leq r}$ denote the shallow VAE layers up to layer $r$, and let $E_{>r}$ denote the remaining layers. For view $c$, the shallow VAE features of the current and previous observations are
\begin{equation}
\label{eq:shallow_features}
    f_q^c
    =
    E_{\leq r}(I_q^c),
    \qquad
    f_{q-1}^c
    =
    E_{\leq r}(I_{q-1}^c).
\end{equation}
The shallow VAE features $f_{q-1}^c$ are cached at the previous policy query. We obtain the observed change map by averaging the channel-wise feature difference:
\begin{equation}
\label{eq:real_change}
    d_{q,\mathrm{obs}}^c(u)
    =
    \frac{1}{C_r}
    \sum_{j=1}^{C_r}
    \left|
        f_q^c(u,j)-f_{q-1}^c(u,j)
    \right|,
\end{equation}
where $u$ indexes spatial positions and $C_r$ is the number of feature channels at layer $r$. The predicted change map is computed from the visual tokens $z_{q-1}^c$ used at query $q-1$ and the predicted future visual tokens $\hat z_{q-1}^{c,\mathrm{future}}$:
\begin{equation}
\label{eq:pred_change}
    d_{q,\mathrm{pred}}^c(v)
    =
    \frac{1}{C_z}
    \sum_{j=1}^{C_z}
    \left|
        \hat z_{q-1}^{c,\mathrm{future}}(v,j)
        -
        z_{q-1}^{c}(v,j)
    \right|,
\end{equation}
where $v$ indexes the spatial positions of the visual tokens and $C_z$ denotes the number of channels in the visual tokens. We align the two maps to the same spatial domain $\Omega$ using adaptive average pooling before comparing them. For a fraction $\rho$ of spatial positions, we define the support of the largest changes as
\begin{equation}
\label{eq:topk_support}
    \mathcal{P}_{\rho}(d)
    =
    \operatorname{TopK}
    \left(
        d,\,
        \rho|\Omega|
    \right).
\end{equation}
The consistency score $o_q^c$ for view $c$ is the intersection-over-union of the observed and predicted change regions. We assign a view-specific consistency threshold $\tau^c$ to each view, because different views exhibit different motion patterns and levels of prediction difficulty. The visual tokens of view $c$ at query $q$ are then
\begin{equation}
\label{eq:vae_route}
    z_q^{c}
    =
    \begin{cases}
        \hat z_{q-1}^{c,\mathrm{future}},
        & o_q^c \geq \tau^c,\\[1mm]
        E_{>r}(f_q^c),
        & \text{otherwise}.
    \end{cases}
\end{equation}
When the predicted tokens are reused, our method does not run the remaining layers $E_{>r}$. To prevent errors from accumulating over long horizons, we reuse visual tokens for at most $N_{\max}$ consecutive policy queries, after which full visual encoding is enabled again.

%% file: sections/sec_experiments.tex
\section{Experiments}
\label{sec:experiments}

In this section, we comprehensively evaluate our method from three perspectives: (1) how its efficiency and task performance compare with existing methods; (2) whether the results generalize across task suites, robotic embodiments, base models and hardware platforms; and (3) how much each acceleration component contributes to the overall gains. To this end, we conduct evaluations in both simulation and real-robot settings.

\subsection{Experimental Setup}

\textbf{Benchmarks.} We first evaluate our method on the widely adopted LIBERO~\citep{liu2023libero}. To further assess generalization, we evaluate in LIBERO-PRO~\citep{zhou2025libero}, which introduces controlled perturbations across multiple factors. We also evaluate in RoboCasa~\citep{nasiriany2024robocasa}, comprising 24 kitchen manipulation tasks with unseen object instances and scene styles.

\textbf{Base Models.} We mainly use Cosmos Policy~\citep{kim2025cosmospolicy} as the base model. We further evaluate our method on additional WAM models (\textit{i.e.}, Fast-WAM~\citep{yuan2026fast} and LingBot-VA~\citep{li2026causal}) and on a representative VLA model, $\pi_{0.5}$~\citep{intelligence2025pi_}.

\textbf{Baselines.} Our baselines include (1) DiCache~\citep{bu2025dicache}, a representative method for video generation models that reuses diffusion residuals; (2) C$^3$ache~\citep{zhao2026c}, specifically designed for WAMs by reducing cross-chunk redundancy; (3) X-Cache~\citep{zeng2026x}, developed for autoregressive world models; and (4) Falcon~\citep{chen2025falcon}, an acceleration method for diffusion policies. Since Falcon relies on unexecuted actions from the previous prediction as references for partial denoising, it can only be evaluated in RoboCasa. We report two variants of our method: \textbf{Ours (Speed)}, which prioritizes maximum acceleration, and \textbf{Ours (Acc.)}, which adopts a more conservative configuration to preserve task success.

\textbf{Implementation Details.} Results reported in the tables are averaged over three random seeds to ensure statistical robustness. All main experiments are conducted on NVIDIA A40 GPUs.

\begin{table*}[!b]
  \centering
  \caption{\textbf{Comparison with representative training-free acceleration methods on the LIBERO~\citep{liu2023libero} and LIBERO-PRO~\citep{zhou2025libero} benchmarks.} \textsuperscript{*} denotes our re-implementation due to unavailable official code. All acceleration methods are adapted to the Cosmos Policy~\citep{kim2025cosmospolicy} architecture for a fair comparison. }
  \label{tab:libero_all}
  \setlength{\tabcolsep}{7pt}
  \renewcommand{\arraystretch}{1.05}
  \resizebox{\textwidth}{!}{
  \begin{tabular}{lcccccccc}
    \toprule
    \multicolumn{1}{l}{\textbf{\textcolor{blue}{LIBERO}}}
    & \multicolumn{4}{c}{\textbf{Success Rate (\%, $\uparrow$) / Speedup ($\uparrow$)}}
    & \multirow{2}{*}{\textbf{\shortstack{Success\\Rate} ($\uparrow$)}}
    & \multirow{2}{*}{\textbf{Speedup ($\uparrow$)}}
    & \multirow{2}{*}{\textbf{Latency ($\downarrow$)}}
    & \multirow{2}{*}{\textbf{FLOPs ($\downarrow$)}} \\
    \cmidrule(lr){2-5}
    \textbf{Method} & \textbf{Goal} & \textbf{Object} & \textbf{Spatial} & \textbf{Long} & & & & \\
    \midrule
    Cosmos Policy & 98.20 / 1.00 & 99.87 / 1.00 & 97.47 / 1.00 & 96.67 / 1.00 & 98.05\% & 1.00$\times$ & 1.4001\,s & 100.00\% \\
    \midrule
    + DiCache & 98.07 / 1.70 & 99.87 / 1.66 & 97.80 / 1.67 & 97.27 / 1.72 & \textbf{98.25\%} & 1.69$\times$ & 0.8273\,s & 48.12\% \\
    + C$^3$ache$^*$ & 97.80 / 2.19 & 99.00 / 2.16 & 97.53 / 2.25 & 94.47 / 2.30 & 97.20\% & 2.24$\times$ & 0.6256\,s & 33.18\% \\
    + X-Cache$^*$ & 97.87 / 1.11 & 99.67 / 1.17 & 97.47 / 1.12 & 97.00 / 1.21 & 98.00\% & 1.17$\times$ & 1.2017\,s & 80.20\% \\
    \rowcolor{rowmint}
    \textbf{+ Ours (Speed)} & 96.73 / 3.89 & 99.07 / 4.08 & 97.60 / 3.87 & 96.27 / 4.07 & 97.42\% & \textbf{4.00$\times$} & \textbf{0.3498\,s} & \textbf{18.39\%} \\
    \rowcolor{rowmint}
    \textbf{+ Ours (Acc.)} & 98.47 / 3.06 & 99.60 / 3.29 & 97.80 / 3.10 & 96.60 / 3.19 & 98.12\% & 3.17$\times$ & 0.4414\,s & 19.19\% \\
    \midrule[\heavyrulewidth]
    \multicolumn{1}{l}{\textbf{\textcolor{blue}{LIBERO-PRO}}}
    & \multicolumn{4}{c}{\textbf{Success Rate (\%, $\uparrow$) / Speedup ($\uparrow$)}}
    & \multirow{2}{*}{\textbf{\shortstack{Success\\Rate} ($\uparrow$)}}
    & \multirow{2}{*}{\textbf{Speedup ($\uparrow$)}}
    & \multirow{2}{*}{\textbf{Latency ($\downarrow$)}}
    & \multirow{2}{*}{\textbf{FLOPs ($\downarrow$)}} \\
    \cmidrule(lr){2-5}
    \textbf{Method} & \textbf{Goal} & \textbf{Object} & \textbf{Spatial} & \textbf{Long} & & & & \\
    \midrule
    Cosmos Policy & 43.67 / 1.00 & 46.13 / 1.00 & 47.40 / 1.00 & 45.33 / 1.00 & 45.63\% & 1.00$\times$ & 1.4044\,s & 100.00\% \\
    \midrule
    + DiCache & 44.13 / 1.67 & 46.27 / 1.66 & 47.73 / 1.63 & 45.07 / 1.73 & \textbf{45.80\%} & 1.68$\times$ & 0.8347\,s & 50.32\% \\
    + C$^3$ache$^*$ & 42.80 / 2.22 & 45.53 / 2.20 & 47.53 / 2.28 & 42.60 / 2.30 & 44.62\% & 2.26$\times$ & 0.6222\,s & 32.86\% \\
    + X-Cache$^*$ & 43.73 / 1.17 & 46.20 / 1.22 & 47.80 / 1.17 & 44.33 / 1.26 & 45.52\% & 1.22$\times$ & 1.1549\,s & 75.55\% \\
    \rowcolor{rowmint}
    \textbf{+ Ours (Speed)} & 44.20 / 3.91 & 45.67 / 3.86 & 47.33 / 3.96 & 44.93 / 3.99 & 45.53\% & \textbf{3.94$\times$} & \textbf{0.3564\,s} & \textbf{18.80\%} \\
    \rowcolor{rowmint}
    \textbf{+ Ours (Acc.)} & 43.67 / 3.05 & 47.00 / 3.08 & 47.60 / 3.06 & 44.60 / 3.11 & 45.72\% & 3.08$\times$ & 0.4558\,s & 19.05\% \\
    \bottomrule
  \end{tabular}
  }
  \vspace{-10pt}
\end{table*}

\subsection{Main Results}

Table~\ref{tab:libero_all} compares our method with Cosmos Policy and representative training-free acceleration baselines. In LIBERO, the original model achieves an average success rate of 98.05\% with a latency of 1.40\,s. DiCache preserves strong task performance through a carefully designed caching strategy, but provides relatively limited gains in inference speed and reductions in computational cost (FLOPs). In contrast, the cross-chunk acceleration methods C$^3$ache and X-Cache still offer limited speedups, while suffering performance degradation. These results show that exploiting redundancy at a single level either provides limited acceleration or incurs noticeable performance degradation. We jointly exploit redundancy across multiple computational levels, yielding a substantially better efficiency-performance trade-off. \textbf{Ours (Speed)} achieves a 4.00$\times$ speedup while preserving comparable task success, whereas \textbf{Ours (Acc.)} provides a more conservative 3.17$\times$ acceleration while slightly exceeding the original performance. Notably, our method reduces FLOPs to only 19\% of the original model, demonstrating substantial computational savings. The acceleration gains are consistent across all four task suites and remain robust on the more challenging LIBERO-PRO benchmark, where our method achieves up to a 3.94$\times$ speedup under distribution shifts. 
\begin{wraptable}{r}{0.5\textwidth}
  \vspace{-20pt}
  \centering
  \caption{\textbf{Evaluation on the RoboCasa benchmark~\citep{nasiriany2024robocasa}.}}
  \label{tab:robocasa}
  \setlength{\tabcolsep}{4pt}
  \renewcommand{\arraystretch}{1.05}
  \resizebox{\linewidth}{!}{
  \begin{tabular}{lcccc}
  \toprule
  \textbf{Method}
  & \textbf{S.R. ($\uparrow$)}
  & \textbf{Speedup ($\uparrow$)}
  & \textbf{Latency ($\downarrow$)}
  & \textbf{FLOPs ($\downarrow$)} \\
  \midrule
  Cosmos Policy & 66.69\% & 1.00$\times$ & 2.2958\,s & 100.00\% \\
  \midrule
  + DiCache & 66.47\% & 1.50$\times$ & 1.5305\,s & 49.99\% \\
  + C$^3$ache$^*$ & 65.00\% & 1.81$\times$ & 1.2653\,s & 33.64\% \\
  + X-Cache$^*$ & 66.37\% & 1.20$\times$ & 1.9202\,s & 72.86\% \\
  + Falcon & 66.47\% & 1.21$\times$ & 1.8968\,s & 74.56\% \\
  \rowcolor{rowmint}
  \textbf{+ Ours} & \textbf{66.94\%} & \textbf{2.10$\times$}
  & \textbf{1.0911\,s} & \textbf{22.50\%} \\
  \bottomrule
  \end{tabular}
  }
  \vspace{-8pt}
\end{wraptable}
We further evaluate our method in RoboCasa to examine its transferability to a different benchmark, where it achieves the best acceleration and task performance, as shown in Table~\ref{tab:robocasa}. Additional experimental analyses, including evaluations across different hardware settings and random seeds, are provided in Appendices~\ref{sec:ana}, \ref{sec:hardware}, and \ref{sec:seeds}. Qualitative results and analyses are provided in Appendix~\ref{sec:qualitative}.

To further evaluate our method with different WAM and VLA base models, we apply its compatible components (\textit{i.e.}, Transformer-level and diffusion-level acceleration) to Fast-WAM~\citep{yuan2026fast}, LingBot-VA~\citep{li2026causal}, and the VLA model $\pi_{0.5}$~\citep{intelligence2025pi_} in LIBERO using publicly released checkpoints. The results are reported in Table~\ref{tab:libero_cross_backbone}. Our method achieves substantial speedups on both WAM backbones while preserving task performance. The speedup on $\pi_{0.5}$ is relatively modest, as a substantial fraction of its latency lies in the VLM backbone. Nevertheless, our method is orthogonal to existing VLA acceleration techniques and can be combined together.
Despite their distinct architectures and inference pipelines, these results demonstrate that our method is not specific to Cosmos Policy and can generalize across different WAM and VLA backbones.

\begin{table*}[h]
\vspace{-10pt}
\centering
\caption{\textbf{Our method with additional WAM and VLA base models on LIBERO~\citep{liu2023libero}.} LingBot-VA can only be evaluated in Long, as only the Long checkpoint is publicly available.}
\label{tab:libero_cross_backbone}
\setlength{\tabcolsep}{7pt}
\renewcommand{\arraystretch}{1.05}
\resizebox{\textwidth}{!}{
\begin{tabular}{lccccccc}
\toprule
\multicolumn{1}{l}{\textbf{LIBERO}}
& \multicolumn{4}{c}{\textbf{Success Rate (\%, $\uparrow$) / Speedup ($\uparrow$)}}
& \multirow{2}{*}{\textbf{\shortstack{Success\\Rate} ($\uparrow$)}}
& \multirow{2}{*}{\textbf{Speedup ($\uparrow$)}}
& \multirow{2}{*}{\textbf{Latency ($\downarrow$)}} \\
\cmidrule(lr){2-5}
\textbf{Method} & \textbf{Goal} & \textbf{Object} & \textbf{Spatial} & \textbf{Long} & & & \\
\midrule
Fast-WAM & 95.87 / 1.00 & 99.40 / 1.00 & 97.13 / 1.00 & 93.93 / 1.00 & \textbf{96.58\%} & 1.00$\times$ & 0.5365\,s\\
\rowcolor{rowmint}
\textbf{+ Ours} & 96.07 / 3.43 & 99.00 / 3.25 & 96.93 / 3.43 & 93.20 / 2.93 & 96.30\% & \textbf{3.17$\times$} & \textbf{0.1693\,s}\\
\midrule
LingBot-VA & -- & -- & -- & 97.40 / 1.00 & 97.40\% & 1.00$\times$ & 13.4105\,s \\
\rowcolor{rowmint}
\textbf{+ Ours} & -- & -- & -- & 98.13 / 2.88 & \textbf{98.13\%} & \textbf{2.88$\times$} & \textbf{4.6558\,s} \\
\midrule
$\pi_{0.5}$ & 97.80 / 1.00 & 98.67 / 1.00 & 98.40 / 1.00 & 92.93 / 1.00 & \textbf{96.95\%} & 1.00$\times$ & 0.1462\,s\\
\rowcolor{rowmint}
\textbf{+ Ours} & 97.87 / 1.29 & 99.07 / 1.30 & 98.47 / 1.33 & 92.27 / 1.24 & 96.92\% & \textbf{1.28$\times$} & \textbf{0.1145\,s}\\
\bottomrule
\end{tabular}
}
\end{table*}

\subsection{Ablation}

\begin{table*}[t]
    \centering
    \scriptsize
    \caption{\textbf{Ablation study on the LIBERO-PRO benchmark~\citep{zhou2025libero}. We evaluate three acceleration components of our method.}}
    \label{tab:libero_ablation}
    \setlength{\tabcolsep}{2pt}
    \renewcommand{\arraystretch}{1.05}
    \begin{tabular}{lcccccccc}
      \toprule
      \multicolumn{1}{l}{\textbf{LIBERO-PRO}}
      & \multicolumn{4}{c}{\textbf{Success Rate (\%, $\uparrow$) / Speedup ($\uparrow$)}}
      & \multirow{2}{*}{\textbf{\shortstack{Success\\Rate} ($\uparrow$)}}
      & \multirow{2}{*}{\textbf{Speedup ($\uparrow$)}}
      & \multirow{2}{*}{\textbf{Latency ($\downarrow$)}}
      & \multirow{2}{*}{\textbf{FLOPs ($\downarrow$)}} \\
      \cmidrule(lr){2-5}
      \textbf{Method} & \textbf{Goal} & \textbf{Object} & \textbf{Spatial} & \textbf{Long}
      & & & & \\
      \midrule
      Cosmos Policy
      & 43.67 / 1.00 & 46.13 / 1.00 & 47.40 / 1.00 & 45.33 / 1.00
      & 45.63\% & 1.00$\times$ & 1.4044\,s & 100.00\% \\
      \midrule
      \rowcolor{rowmint}
      \textbf{Ours (Speed)}
      & 44.20 / 3.91 & 45.67 / 3.86 & 47.33 / 3.96 & 44.93 / 3.99
      & 45.53\% & \textbf{3.94$\times$} & \textbf{0.3564\,s}
      & \textbf{18.80\%} \\
      \textit{w/o} Action-Aware Step Skipping
      & 43.67 / 1.18 & 44.67 / 1.18 & 47.00 / 1.18 & 44.47 / 1.21
      & 44.95\% & 1.19$\times$ & 1.1803\,s & 92.81\% \\
      \textit{w/o} Attention Skipping
      & 43.80 / 3.64 & 45.67 / 3.59 & 47.80 / 3.69 & 43.93 / 3.69
      & 45.30\% & 3.66$\times$ & 0.3841\,s & 20.19\% \\
      \textit{w/o} Visual Token Reuse
      & 43.60 / 3.07 & 46.87 / 3.01 & 47.53 / 3.08 & 44.67 / 3.07
      & \textbf{45.67\%} & 3.06$\times$ & 0.4595\,s & 19.05\% \\
      \bottomrule
    \end{tabular}
    \vspace{-12pt}
\end{table*}

\textbf{Contribution of Each Acceleration Component.}
Table~\ref{tab:libero_ablation} evaluates all three components of our method on LIBERO-PRO. Action-Aware Step Skipping provides the largest acceleration gain, indicating substantial redundancy across diffusion denoising steps. Visual Token Reuse further contributes considerable speedup by reducing the increasingly dominant VAE encoding cost, while Attention Skipping provides an additional efficiency gain by eliminating redundant computation within the Transformer. Importantly, removing any individual module has only a minor effect on task success, suggesting that these modules exploit complementary sources of redundancy at different computational levels. Their joint application achieves the best overall efficiency-performance trade-off, reaching a 3.94$\times$ speedup with negligible performance degradation.

\begin{table*}[t]
    \centering
    \caption{\textbf{Effectiveness of visual reuse verification on LIBERO-PRO.}}
    \label{tab:verifier}
    \setlength{\tabcolsep}{7pt}
    \renewcommand{\arraystretch}{1.05}
    \resizebox{\textwidth}{!}{
    \begin{tabular}{lcccccccc}
        \toprule
        \multicolumn{1}{l}{\textbf{LIBERO-PRO}}
        & \multicolumn{4}{c}{\textbf{Success Rate (\%, $\uparrow$) / Speedup ($\uparrow$)}}
        & \multirow{2}{*}{\textbf{\shortstack{Success\\Rate} ($\uparrow$)}}
        & \multirow{2}{*}{\textbf{Speedup ($\uparrow$)}}
        & \multirow{2}{*}{\textbf{Latency ($\downarrow$)}}
        & \multirow{2}{*}{\textbf{\shortstack{Skip\\Rate}}} \\
        \cmidrule(lr){2-5}
        \textbf{Method} & \textbf{Goal} & \textbf{Object} & \textbf{Spatial} & \textbf{Long} & & & & \\
        \midrule
        Cosmos Policy
        & 43.67 / 1.00 & 46.13 / 1.00 & 47.40 / 1.00 & 45.33 / 1.00
        & 45.63\% & 1.00$\times$ & 1.4043\,s & 0.00\% \\
        \midrule
        Random Reuse
        & 42.87 / 4.09 & 45.80 / 4.11 & 46.80 / 4.09 & 44.00 / 4.20
        & 44.87\% & 4.14$\times$ & 0.3395\,s & 56.44\% \\
        Periodic Refresh
        & 43.13 / 4.25 & 45.07 / 4.32 & 46.53 / 4.29 & 44.60 / 4.38
        & 44.83\% & 4.32$\times$ & 0.3249\,s & 64.99\% \\
        Unconditional Reuse
        & 40.67 / 5.61 & 40.07 / 5.63 & 38.53 / 5.57 & 32.87 / 5.88
        & 38.03\% & 5.72$\times$ & 0.2457\,s & 94.05\% \\
        \rowcolor{rowmint}
        \textbf{Ours (Speed)}
        & 44.20 / 3.91 & 45.67 / 3.86 & 47.33 / 3.96 & 44.93 / 3.99
        & \textbf{45.53\%} & 3.94$\times$ & 0.3564\,s & 56.25\% \\
        \bottomrule
    \end{tabular}
    }

    \centering
    \caption{\textbf{Sensitivity Analysis of the VAE Feature Layer in LIBERO-PRO.} Layer-1 achieves the best task performance and inference efficiency among the evaluated feature layers.}
    \label{tab:vae_layer_sensitivity}
    \setlength{\tabcolsep}{6pt}
    \renewcommand{\arraystretch}{1.05}
    \resizebox{\textwidth}{!}{
    \begin{tabular}{lcccccccc}
    \toprule
    \multicolumn{1}{l}{\textbf{LIBERO-PRO}}
    & \multicolumn{4}{c}{\textbf{Success Rate (\%, $\uparrow$) / Speedup ($\uparrow$)}}
    & \multirow{2}{*}{\textbf{\shortstack{Success\\Rate} ($\uparrow$)}}
    & \multirow{2}{*}{\textbf{Speedup ($\uparrow$)}}
    & \multirow{2}{*}{\textbf{Latency ($\downarrow$)}}
    & \multirow{2}{*}{\textbf{\shortstack{VAE Skip\\Rate}}} \\
    \cmidrule(lr){2-5}
    \textbf{Method} & \textbf{Goal} & \textbf{Object} & \textbf{Spatial} & \textbf{Long} & & & & \\
    \midrule
    Cosmos Policy & 43.67 / 1.00 & 46.13 / 1.00 & 47.40 / 1.00 & 45.33 / 1.00 & 45.63\% & 1.00$\times$ & 1.4043\,s & -- \\
    \midrule
    Layer-0 & 44.13 / 3.91 & 45.07 / 3.86 & 46.93 / 3.91 & 44.73 / 3.98 & 45.22\% & 3.93$\times$ & 0.3574\,s & 56.20\% \\
    \rowcolor{rowmint}
    \textbf{Layer-1} & \textbf{44.20 / 3.91} & \textbf{45.67 / 3.86} & \textbf{47.33 / 3.96} & \textbf{44.93 / 3.99} & \textbf{45.53\%} & \textbf{3.94$\times$} & \textbf{0.3564\,s} &
    56.25\% \\
    Layer-2 & 43.67 / 3.85 & 45.27 / 3.79 & 46.87 / 3.92 & 45.33 / 3.93 & 45.28\% & 3.88$\times$ & 0.3621\,s & 56.41\% \\
    Layer-3 & 43.73 / 3.83 & 45.40 / 3.78 & 47.20 / 3.90 & 45.33 / 3.91 & 45.42\% & 3.86$\times$ & 0.3638\,s & 56.44\% \\
    \bottomrule
    \end{tabular}
    }
\vspace{-12pt}
\end{table*}

\textbf{Further Analysis of Visual Token Reuse.} To assess whether verification provides benefits beyond forced refreshes, we compare our method with random reuse, periodic refresh, and unconditional reuse on LIBERO-PRO. As shown in Table~\ref{tab:verifier}, we improves success over random reuse at closely matched skip rates. It also outperforms periodic refresh by 0.70\% while reusing visual tokens less frequently. Unconditional reuse achieves the highest skip rate but substantially reduces task success. These results suggest that verification provides benefits beyond refresh frequency alone.

We further analyze Visual Token Reuse by varying the depth of the shallow VAE layer probe. A shallower probe incurs less verification overhead but provides lower-level visual features, whereas deeper layers capture more abstract visual representations and tend to yield slightly higher skip rates, at the cost of additional encoding computation. 
The results are reported in Table~\ref{tab:vae_layer_sensitivity}. Overall, Layer-1 achieves the best balance between reuse rate and verification overhead while maintaining a high task success rate, and is therefore adopted as the default shallow probe. Nevertheless, our method maintains comparable task performance and substantial acceleration across different probe depths, indicating that it does not rely on a carefully tuned probe layer to achieve a favorable efficiency-performance trade-off. Additional latency analyses and parameter sensitivity analyses are provided in Appendices~\ref{sec:latency} and \ref{sec:sensitivity}.

\subsection{Evaluation on the real-robot task}
\label{sec:realworld}

\begin{figure}[H]
\vspace{-8pt}
\centering
\includegraphics[width=\linewidth]{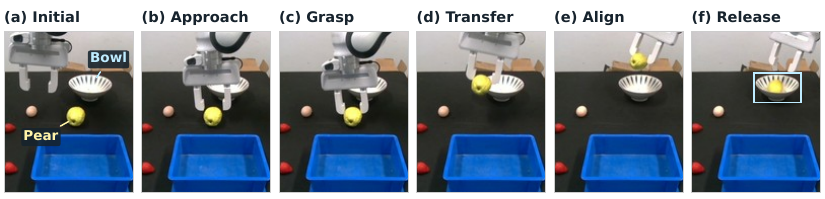}
\vspace{-20pt}
\caption{\textbf{The real-robot pear-to-bowl task.} A recorded demonstration, shown from left to right.}
\vspace{-12pt}
\label{fig:realworld_sequence}
\end{figure}

\begin{wraptable}{r}{0.46\textwidth}
\vspace{-8pt}
\centering
\caption{\textbf{Evaluation on the real-robot task on a Franka robot.}}
\label{tab:realworld}
\setlength{\tabcolsep}{4pt}
\renewcommand{\arraystretch}{1.05}
\resizebox{\linewidth}{!}{
\begin{tabular}{lccc}
\toprule
\textbf{Method}
& \textbf{S.R. ($\uparrow$)}
& \textbf{Speedup ($\uparrow$)}
& \textbf{Latency ($\downarrow$)} \\
\midrule
Cosmos Policy & 90.00\% & 1.00$\times$ & 0.6949\,s \\
\rowcolor{rowmint}
\textbf{+ Ours} & \textbf{95.00\%} & \textbf{2.65$\times$} & \textbf{0.2625\,s} \\
\midrule
$\pi_{0.5}$ & 25.00\% & 1.00$\times$ & 0.1463\,s \\
\rowcolor{rowmint}
\textbf{+ Ours} & \textbf{25.00\%} & \textbf{1.21$\times$} & \textbf{0.1209\,s} \\
\bottomrule
\end{tabular}
}
\end{wraptable}
We evaluate our method across 40 trials on the real-robot pick-and-place task on Franka Panda, using an NVIDIA H20 for Cosmos Policy and an RTX 3090 for $\pi_{0.5}$. The task is to pick up the pear and place it in the bowl, as illustrated in Figure~\ref{fig:realworld_sequence}. As shown in Table~\ref{tab:realworld}, our method effectively accelerates inference while preserving the success rate. These results further demonstrate the practical applicability of our method.

%% file: sections/sec_conclusion.tex
\section{Conclusion}
\label{sec:conclusion}

We presented MARS, a training-free and plug-and-play framework for accelerating World Action Model inference. 
Our analysis reveals that WAMs contain substantial but highly non-uniform redundancy across Transformer computation, diffusion denoising, and consecutive policy queries, and that the safe removal of this redundancy depends on control sensitivity. 
Guided by these observations, MARS selectively reduces low-impact attention computation, adapts diffusion acceleration to motion granularity, and reuses predicted future visual representations with online verification. 
Across LIBERO, LIBERO-PRO, and RoboCasa, MARS achieves substantial end-to-end speedups while preserving task performance, and remains effective across different model backbones and hardware platforms. 
These results suggest that holistic, control-aware redundancy reduction is a promising direction for making generative world-action policies more practical for real-time robotic control.

%% file: sections/appendix.tex
\newpage
\appendix

\section{Experimental Settings}
\label{sec:set}
\subsection{Simulation and Real-World Environments}
\label{sec:environment}
We evaluate our method on three simulation benchmarks and a real-world environment covering diverse robotic manipulation tasks and generalization settings:
\begin{itemize}
    \item \textbf{LIBERO~\citep{liu2023libero}:}
    LIBERO is a widely used benchmark for evaluating generalizable robot manipulation policies. We evaluate on its four task suites, including LIBERO-Spatial, LIBERO-Object, LIBERO-Goal, and LIBERO-Long, which assess spatial reasoning, object interaction, goal-conditioned manipulation, and long-horizon task execution, respectively. In total, we conduct 2,000 evaluation trajectories to assess model robustness.

    \item \textbf{LIBERO-PRO~\citep{zhou2025libero}:}
    LIBERO-PRO extends LIBERO with controlled distribution shifts across five core generalization dimensions, including Object, which varies object appearance, color, and scale; Position, which changes object locations within feasible spatial regions; Semantic, which paraphrases language instructions; Task, which modifies task logic and target states; and Environment, which introduces different scene layouts and environments. We conduct 2,000 evaluation trajectories to examine whether acceleration methods remain robust under these challenging distribution shifts.

    \item \textbf{RoboCasa~\citep{nasiriany2024robocasa}:}
    RoboCasa provides diverse household manipulation tasks in realistic kitchen environments. We evaluate 24 tasks using a Franka Emika Panda robot. For each task, the success rate is measured over 50 trials across five evaluation scenes with different floor plans and styles, resulting in 1,200 trials. All evaluation objects are unseen instances, and two of the five scenes for each task additionally use styles not observed during training, providing a challenging test of visual and environmental generalization.

    \item \textbf{Real-World Environment:}
    We further evaluate our method on real-world robotic manipulation tasks using the Franka platform. These experiments assess whether our method can provide practical inference acceleration while preserving closed-loop performance under real-world visual observations and physical interactions.
    
\end{itemize}

\subsection{Base Models}
\label{sec:target}
We evaluate our method across multiple WAM and VLA architectures to examine its architectural generality:
\begin{itemize}
    \item \textbf{Cosmos Policy~\citep{kim2025cosmospolicy}:}
    Cosmos Policy serves as our primary experimental backbone. 
    It adapts the pretrained Cosmos-Predict2 video diffusion model to jointly generate robot actions, future visual observations, proprioceptive states, and state values within a unified generative process.

    \item \textbf{Fast-WAM~\citep{yuan2026fast}:}
    Fast-WAM retains video modeling during training while removing explicit future visual generation at inference time. 
    We use it to evaluate whether the compatible components of our method generalize to WAMs without test-time future imagination.

    \item \textbf{LingBot-VA~\citep{li2026causal}:}
    LingBot-VA is an autoregressive diffusion-based WAM that jointly performs future-frame prediction and policy execution. 
    Its distinct autoregressive inference pipeline provides an additional setting for evaluating our method.

    \item \textbf{$\pi_{0.5}$~\citep{intelligence2025pi_}:}
    $\pi_{0.5}$ is a general-purpose VLA model designed for broad open-world manipulation generalization. 
    We apply the compatible acceleration components of our method to examine its transferability beyond WAM architectures.
\end{itemize}

\subsection{Baselines}
\label{sec:baselines}
We compare our method with representative training-free inference acceleration methods. Methods marked with * are our implementations, as official code is unavailable. All baselines are adapted to the same Cosmos Policy backbone for a fair comparison:
\begin{itemize}
    \item \textbf{DiCache~\citep{bu2025dicache}:}
    DiCache is a training-free diffusion acceleration method that exploits redundancy across consecutive denoising steps. 
    It estimates intermediate feature changes and reuses cached residuals when the variation is sufficiently small.

    \item \textbf{C$^3$ache~\citep{zhao2026c}:}
    C$^3$ache exploits redundancy across consecutive inference chunks in WAMs. 
    It caches and reuses DiT residuals at corresponding denoising steps across action chunks to reduce repeated computation.

    \item \textbf{X-Cache~\citep{zeng2026x}:}
    X-Cache is a training-free cross-chunk caching method that performs fine-grained block-level reuse. 
    It employs structure- and action-aware fingerprints to determine whether individual Transformer blocks should recompute or reuse residuals.

    \item \textbf{Falcon~\citep{chen2025falcon}:}
    Falcon is a training-free diffusion policy acceleration method that exploits temporal dependencies across control steps. It reuses historical partially denoised actions as the initialization for current denoising, using one-step estimation and a thresholding mechanism to select suitable starting points and reduce sampling steps.
\end{itemize}

\section{Analysis of Experimental Results}
\label{sec:ana}
Table~\ref{tab:libero_all} compares our method with representative training-free acceleration methods in LIBERO and LIBERO-PRO. The results reveal three main observations:

\noindent\textbf{(1) Carefully controlled caching preserves accuracy but provides limited acceleration.}
DiCache preserves task performance well, achieving 98.25\% and 45.80\% success rates in LIBERO and LIBERO-PRO, respectively, even slightly outperforming the original policy. However, it provides only 1.69$\times$ and 1.68$\times$ speedups. This suggests that diffusion caching strategies originally designed for video generation are overly conservative for WAMs, leaving substantial computational redundancy unexploited.

\noindent\textbf{(2) Aggressive cross-chunk reuse can impair long-horizon task execution.}
C$^3$ache and X-Cache exhibit larger performance degradation, particularly on the Long suite, indicating that approximation errors introduced by cross-chunk computation reuse can accumulate over consecutive action chunks. This highlights the sensitivity of WAMs to aggressive computation reuse across temporally dependent action chunks.

\noindent\textbf{(3) Multi-level redundancy elimination yields a better efficiency-performance trade-off.}
Our method jointly exploits redundancy across Transformer components, denoising steps, and consecutive action chunks. \textbf{Ours (Speed)} achieves 4.00$\times$/3.94$\times$ speedups in LIBERO/LIBERO-PRO, while \textbf{Ours (Acc.)} achieves 3.17$\times$/3.08$\times$ acceleration and preserves, and even slightly improves the success rate of the original policy. A similar trend is observed on the RoboCasa benchmark, where our method provides substantial acceleration while maintaining task performance. These results demonstrate that coordinated multi-level acceleration provides a more favorable efficiency-performance trade-off than exploiting a single source of redundancy.

\begin{table*}[h]
  \centering
  \caption{\textbf{Comparison of our method across different GPUs in LIBERO.}}
  \label{tab:libero_device}
  \setlength{\tabcolsep}{7pt}
  \renewcommand{\arraystretch}{1.05}
  \resizebox{\textwidth}{!}{
  \begin{tabular}{lccccccc}
    \toprule
    \multicolumn{1}{l}{\textbf{LIBERO}}
    & \multicolumn{4}{c}{\textbf{Success Rate (\%, $\uparrow$) / Speedup ($\uparrow$)}}
    & \multirow{2}{*}{\textbf{\shortstack{Success\\Rate} ($\uparrow$)}}
    & \multirow{2}{*}{\textbf{Speedup ($\uparrow$)}}
    & \multirow{2}{*}{\textbf{Latency ($\downarrow$)}} \\
    \cmidrule(lr){2-5}
    \textbf{Method} & \textbf{Goal} & \textbf{Object} & \textbf{Spatial} & \textbf{Long} & & & \\
    \midrule
    \multicolumn{8}{c}{\textbf{\textcolor{blue}{A40}}} \\
    \midrule
    Cosmos Policy & 98.20 / 1.00 & 99.87 / 1.00 & 97.47 / 1.00 & 96.67 / 1.00 & 98.05\% & 1.00$\times$ & 1.4001\,s \\
    \rowcolor{rowmint}\textbf{+ Ours (Speed)} & 96.73 / 3.89 & 99.07 / 4.08 & 97.60 / 3.87 & 96.27 / 4.07 & 97.42\% & \textbf{4.00$\times$} & \textbf{0.3498\,s} \\
    \midrule
    \multicolumn{8}{c}{\textbf{\textcolor{blue}{RTX 3090}}} \\
    \midrule
    Cosmos Policy & 97.67 / 1.00 & 100.00 / 1.00 & 97.53 / 1.00 & 97.40 / 1.00 & 98.15\% & 1.00$\times$ & 1.4181\,s \\
    \rowcolor{rowmint}\textbf{+ Ours (Speed)} & 97.07 / 3.97 & 99.53 / 4.16 & 98.13 / 3.93 & 96.13 / 4.10 & 97.72\% & \textbf{4.06$\times$} & \textbf{0.3496\,s} \\
    \midrule
    \multicolumn{8}{c}{\textbf{\textcolor{blue}{RTX 5090}}} \\
    \midrule
    Cosmos Policy & 98.27 / 1.00 & 100.00 / 1.00 & 97.27 / 1.00 & 97.60 / 1.00 & 98.28\% & 1.00$\times$ & 0.5698\,s \\
    \rowcolor{rowmint}\textbf{+ Ours (Speed)} & 97.13 / 3.66 & 99.33 / 3.82 & 98.00 / 3.63 & 95.60 / 3.73 & 97.52\% & \textbf{3.72$\times$} & \textbf{0.1532\,s} \\
    \bottomrule
  \end{tabular}
  }
\end{table*}

\section{Evaluation Across Hardware Platforms}
\label{sec:hardware}
To assess the hardware generality of our method, we benchmark its inference efficiency on three GPU platforms: NVIDIA GeForce RTX 3090, NVIDIA GeForce RTX 5090, and NVIDIA A40. As shown in Table~\ref{tab:libero_device}, our method consistently reduces end-to-end latency across all devices on the LIBERO benchmark. Although the absolute latency and observed speedup vary with hardware architecture, memory bandwidth, and kernel efficiency, the acceleration ratio remains consistently around $4\times$. These results demonstrate that our method is not tailored to a specific accelerator and can provide practical efficiency gains across both consumer-grade and datacenter GPUs.

\section{Results Across Random Seeds}
\label{sec:seeds}
To examine the robustness of our results to evaluation randomness, we report the complete results in the LIBERO simulation using three random seeds (195, 196, and 197) in Table~\ref{tab:libero_multiseed}. The results reported in the main-text tables are averaged over these three runs. Across different seeds, our method exhibits highly consistent performance and acceleration. In particular, Ours (Speed) achieves average success rates of 97.45\%, 97.30\%, and 97.50\%, together with speedups of $4.04\times$, $3.99\times$, and $3.98\times$, respectively. These results demonstrate that the effectiveness of our method is robust to different random seeds, with only minor variations in task success rate and inference efficiency.

\section{Latency Analysis}
\label{sec:latency}
\begin{figure}[t]
    % \vspace{-8pt}
    \centering
    \includegraphics[width=\textwidth]{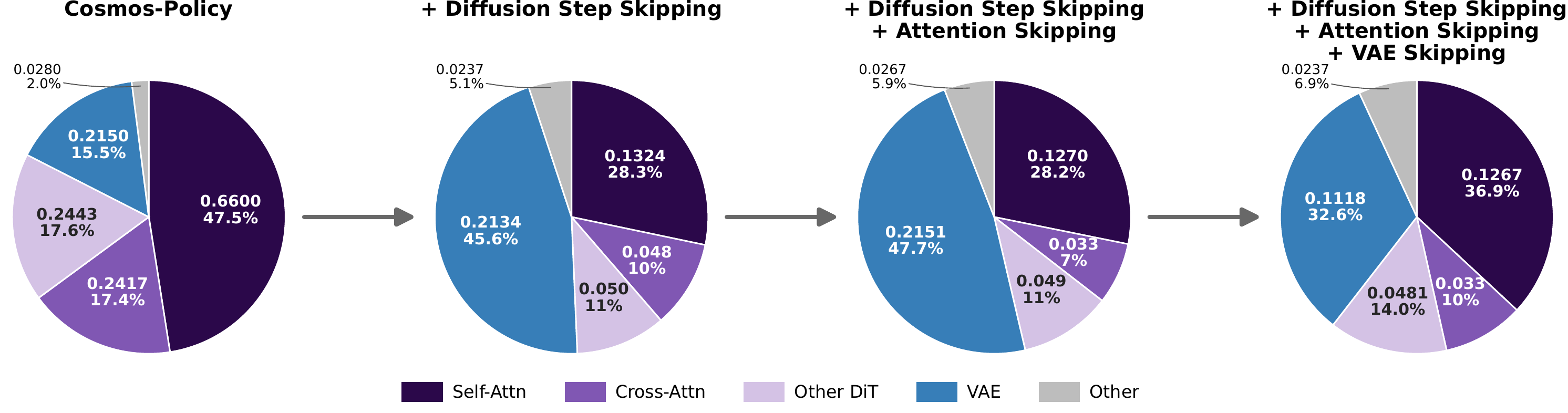}
    % \vspace{-7mm}
    \caption{\textbf{Analysis of the latency.} Component-wise latency breakdown and the resulting shift in computational bottlenecks as different acceleration components are progressively introduced.}
    \label{fig:analysis}
    % \vspace{-4mm}
\end{figure}

Figure~\ref{fig:analysis} shows how the computational bottleneck shifts as the three acceleration modules are progressively introduced in LIBERO. In Cosmos Policy, DiT computation dominates the overall latency. Action-Aware Step Skipping substantially reduces repeated DiT computation, lowering the total measured latency from 1.40\,s to 0.47\,s. Attention Skipping further reduces Transformer cost by eliminating low-impact attention computation. Consequently, VAE encoding becomes the dominant remaining bottleneck, accounting for 47.7\% of the latency in the accelerated pipeline. Finally, Visual Token Reuse reduces VAE encoding latency, achieving nearly a $4\times$ end-to-end speedup. This progressive shift in the latency composition validates the effectiveness of multi-level acceleration for WAM inference.

\begin{table*}[!b]
  \vspace{-6pt}
  \centering
  \caption{\textbf{Results in LIBERO with different random seeds.}}
  \label{tab:libero_multiseed}
  \setlength{\tabcolsep}{7pt}
  \renewcommand{\arraystretch}{1.05}
  \resizebox{\textwidth}{!}{
  \begin{tabular}{lccccccc}
    \toprule
    \multicolumn{1}{l}{\textbf{LIBERO}}
    & \multicolumn{4}{c}{\textbf{Success Rate (\%, $\uparrow$) / Speedup ($\uparrow$)}}
    & \multirow{2}{*}{\textbf{\shortstack{Success\\Rate} ($\uparrow$)}}
    & \multirow{2}{*}{\textbf{Speedup ($\uparrow$)}}
    & \multirow{2}{*}{\textbf{Latency ($\downarrow$)}} \\
    \cmidrule(lr){2-5}
    \textbf{Method} & \textbf{Goal} & \textbf{Object} & \textbf{Spatial} & \textbf{Long} & & & \\
    \midrule

    \multicolumn{8}{c}{\textbf{\textcolor{blue}{Seed 195}}} \\
    \midrule
    Cosmos Policy & 98.20 / 1.00 & 99.80 / 1.00 & 98.00 / 1.00 & 96.00 / 1.00 & 98.00\% & 1.00$\times$ & 1.4194\,s \\
    + DiCache & 98.20 / 1.69 & 100.00 / 1.59 & 98.60 / 1.65 & 96.80 / 1.60 & \textbf{98.40\%} & 1.62$\times$ & 0.8737\,s \\
    + C$^3$ache$*$ & 98.40 / 2.20 & 99.00 / 2.18 & 97.00 / 2.26 & 95.20 / 2.39 & 97.40\% & 2.28$\times$ & 0.6215\,s \\
    + X-Cache$^*$ & 97.80 / 1.10 & 99.40 / 1.17 & 97.80 / 1.12 & 95.80 / 1.24 & 97.70\% & 1.18$\times$ & 1.2036\,s \\
    \rowcolor{rowmint}
    \textbf{+ Ours (Speed)} & 96.20 / 3.86 & 99.60 / 4.08 & 97.80 / 3.86 & 96.20 / 4.19 & 97.45\% & \textbf{4.04$\times$} & \textbf{0.3511\,s} \\
    \rowcolor{rowmint}
    \textbf{+ Ours (Acc.)} & 99.00 / 3.06 & 99.40 / 3.29 & 98.20 / 3.10 & 96.60 / 3.29 & 98.30\% & 3.21$\times$ & 0.4417\,s \\

    \midrule
    \multicolumn{8}{c}{\textbf{\textcolor{blue}{Seed 196}}} \\
    \midrule
    Cosmos Policy & 99.00 / 1.00 & 100.00 / 1.00 & 97.20 / 1.00 & 96.60 / 1.00 & \textbf{98.20\%} & 1.00$\times$ & 1.3935\,s \\
    + DiCache & 98.40 / 1.71 & 100.00 / 1.70 & 97.20 / 1.68 & 97.20 / 1.79 & \textbf{98.20\%} & 1.74$\times$ & 0.8030\,s \\
    + C$^3$ache$*$ & 98.00 / 2.18 & 98.80 / 2.16 & 97.80 / 2.25 & 93.20 / 2.26 & 96.95\% & 2.22$\times$ & 0.6279\,s \\
    + X-Cache$^*$ & 98.20 / 1.10 & 99.80 / 1.17 & 97.00 / 1.13 & 97.00 / 1.19 & 98.00\% & 1.16$\times$ & 1.2039\,s \\
    \rowcolor{rowmint}
    \textbf{+ Ours (Speed)} & 97.20 / 3.91 & 98.80 / 4.08 & 97.60 / 3.89 & 95.60 / 4.01 & 97.30\% & \textbf{3.99$\times$} & \textbf{0.3496\,s} \\
    \rowcolor{rowmint}
    \textbf{+ Ours (Acc.)} & 98.20 / 3.07 & 99.40 / 3.32 & 97.60 / 3.10 & 95.80 / 3.16 & 97.75\% & 3.16$\times$ & 0.4403\,s \\

    \midrule
    \multicolumn{8}{c}{\textbf{\textcolor{blue}{Seed 197}}} \\
    \midrule
    Cosmos Policy & 97.40 / 1.00 & 99.80 / 1.00 & 97.20 / 1.00 & 97.40 / 1.00 & 97.95\% & 1.00$\times$ & 1.3873\,s \\
    + DiCache & 97.60 / 1.70 & 99.60 / 1.68 & 97.60 / 1.66 & 97.80 / 1.78 & 98.15\% & 1.72$\times$ & 0.8052\,s \\
    + C$^3$ache$*$ & 97.00 / 2.18 & 99.20 / 2.15 & 97.80 / 2.24 & 95.00 / 2.25 & 97.25\% & 2.21$\times$ & 0.6273\,s \\
    + X-Cache$^*$ & 97.60 / 1.11 & 99.80 / 1.17 & 97.60 / 1.12 & 98.20 / 1.19 & 98.30\% & 1.16$\times$ & 1.1976\,s \\
    \rowcolor{rowmint}
    \textbf{+ Ours (Speed)} & 96.80 / 3.90 & 98.80 / 4.07 & 97.40 / 3.86 & 97.00 / 4.02 & 97.50\% & \textbf{3.98$\times$} & \textbf{0.3486\,s} \\
    \rowcolor{rowmint}
    \textbf{+ Ours (Acc.)} & 98.20 / 3.06 & 100.00 / 3.27 & 97.60 / 3.09 & 97.40 / 3.12 & \textbf{98.30\%} & 3.14$\times$ & 0.4423\,s \\

    \bottomrule
  \end{tabular}
  }
  \vspace{-5pt}
\end{table*}

\begin{table*}[t]
    % \vspace{-8pt}
    \centering
    \caption{\textbf{Sensitivity Analysis of the Action Routing Threshold $\tau_{\mathrm{act}}$ in LIBERO-PRO.} Our method maintains comparable task performance and substantial acceleration across a broad range of action threshold values.}
    \label{tab:key_parameter_sensitivity}
    \setlength{\tabcolsep}{7pt}
    \renewcommand{\arraystretch}{1.05}
    \resizebox{\textwidth}{!}{
    \begin{tabular}{lccccccc}
      \toprule
      \multicolumn{1}{l}{\textbf{LIBERO-PRO}} & \multicolumn{4}{c}{\textbf{Success Rate (\%, $\uparrow$) / Speedup ($\uparrow$)}} & \multirow{2}{*}{\textbf{\shortstack{Success\\Rate} ($\uparrow$)}} & \multirow{2}
      {*}{\textbf{Speedup ($\uparrow$)}} & \multirow{2}{*}{\textbf{Latency ($\downarrow$)}} \\
      \cmidrule(lr){2-5}
      \textbf{Method} & \textbf{Goal} & \textbf{Object} & \textbf{Spatial} & \textbf{Long} & & & \\
      \midrule
      Cosmos Policy & 43.67 / 1.00 & 46.13 / 1.00 & 47.40 / 1.00 & 45.33 / 1.00 & 45.63\% & 1.00$\times$ & 1.4043\,s \\
      \midrule
      Ours-1 (0.0) & 44.00 / 4.01 & 45.47 / 3.95 & 46.80 / 4.02 & 44.27 / 4.10 & 45.13\% & \textbf{4.03$\times$} & \textbf{0.3483\,s} \\
      Ours-2 (0.2) & 43.73 / 3.93 & 45.60 / 3.84 & 46.80 / 3.95 & 43.80 / 4.04 & 44.98\% & 3.96$\times$ & 0.3550\,s \\
      \rowcolor{rowmint}
      \textbf{Ours-3 (0.4)} & \textbf{44.20 / 3.91} & \textbf{45.67 / 3.86} & \textbf{47.33 / 3.96} & \textbf{44.93 / 3.99} & \textbf{45.53\%} & 3.94$\times$ & 0.3564\,s \\
      Ours-4 (0.6) & 43.93 / 3.90 & 45.47 / 3.86 & 47.27 / 3.95 & 44.40 / 3.99 & 45.27\% & 3.94$\times$ & 0.3567\,s \\
      Ours-5 (0.8) & 43.00 / 3.87 & 45.60 / 3.82 & 47.47 / 3.91 & 44.27 / 3.96 & 45.08\% & 3.90$\times$ & 0.3601\,s \\
      Ours-6 (1.0) & 43.73 / 3.85 & 45.53 / 3.78 & 47.73 / 3.88 & 43.67 / 3.94 & 45.17\% & 3.87$\times$ & 0.3625\,s \\
      Ours-7 ($+\infty$) & 43.60 / 3.89 & 45.53 / 3.82 & 47.87 / 3.89 & 43.60 / 3.98 & 45.15\% & 3.84$\times$ & 0.3655\,s \\
      \bottomrule
    \end{tabular}
    }
    % \vspace{-10pt}
\end{table*}

\section{Sensitivity Analysis of the Action Routing Threshold}
\label{sec:sensitivity}
The action routing threshold $\tau_{\mathrm{act}}$ controls the trade-off between aggressive one-step generation and the more conservative Multi-Step Solver with Residual Reuse. To examine the sensitivity of our method to this hyperparameter, we vary $\tau_{\mathrm{act}}$ over $\{0.0, 0.2, 0.4, 0.6, 0.8, 1.0, +\infty\}$ and evaluate both task success rate and end-to-end inference speedup, as reported in Table~\ref{tab:key_parameter_sensitivity}. A smaller threshold routes more action chunks to direct one-step generation, yielding higher acceleration but potentially reducing robustness for fine-grained manipulation, whereas a larger threshold retains Multi-Step Solver with Residual Reuse for more queries and therefore favors task performance at the cost of additional computation. With the default threshold $\tau_{\mathrm{act}}=0.4$, routing improves success rates across all four suites compared with always-one-step sampling ($\tau_{\mathrm{act}}=0$). It also achieves a higher average success rate with lower latency than always using the multi-step solver with residual reuse ($\tau_{\mathrm{act}}=+\infty$). Across the tested range of $\tau_{\mathrm{act}}$, our method consistently maintains comparable task performance and substantial acceleration, demonstrating that its effectiveness is not sensitive to a particular threshold choice and does not rely on specific hyperparameter tuning.

\section{Qualitative Results and Analyses}
\label{sec:qualitative}

To provide a deeper understanding of our proposed framework, we compare Ours (Acc.) with the original Cosmos Policy through qualitative visualizations in both LIBERO and LIBERO-PRO. We analyze 12,000 paired episodes across three random seeds and identify 186 cases in which Ours succeeds while Cosmos Policy fails, compared with 177 cases exhibiting the opposite outcome. Although these changes largely offset each other, Ours still achieves a slight overall improvement in success rate.

\begin{figure}[H]
    \centering
    \vspace{-15pt}
    \includegraphics[width=\textwidth]{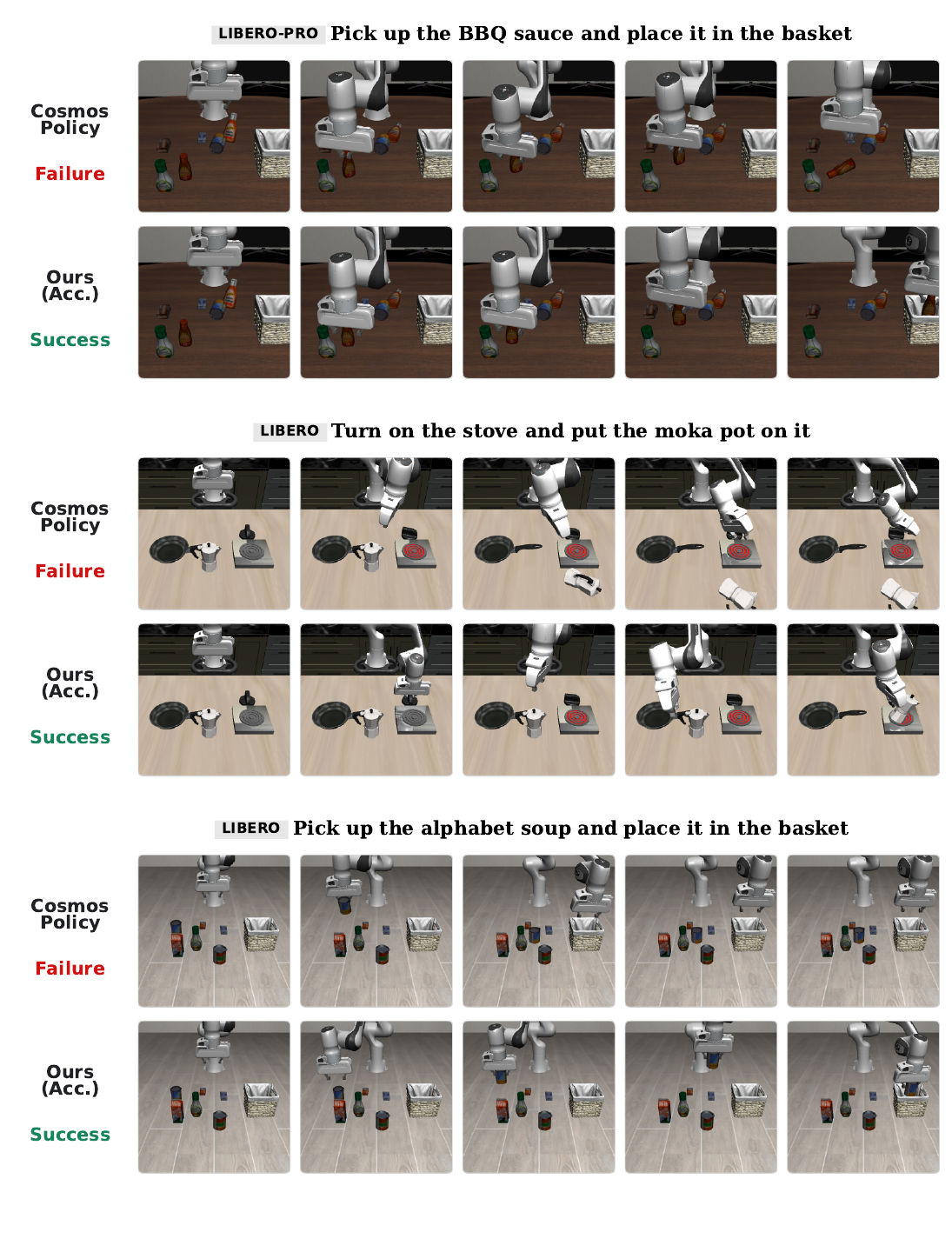}
    \vspace{-45pt}
    \caption{\textbf{Qualitative examples where our method improves task completion.}
    Representative comparisons in LIBERO and LIBERO-PRO where Ours (Acc.) succeeds while the original Cosmos Policy fails.}
    \label{fig:visual1}
\end{figure}

Ours more often recovers trajectories involving direct object transport, where Cosmos Policy may contact or grasp the target but fail to deliver it to the desired location. As shown in Figure~\ref{fig:visual1}, the qualitative comparisons include representative tasks where Ours succeeds while the original policy fails. We hypothesize that using direct one-step denoising more frequently during coarse-grained manipulation stages can facilitate task completion. This advantage is particularly evident across object pick-and-place tasks in LIBERO-PRO, where Ours yields a net gain of 17 successful episodes over the original Cosmos Policy. In the shown cases, Cosmos Policy prematurely opens the gripper during large motions, whereas Ours maintains a stable grasp until placement.

\begin{figure}[H]
    \centering
    \vspace{-15pt}
    \includegraphics[width=\textwidth]{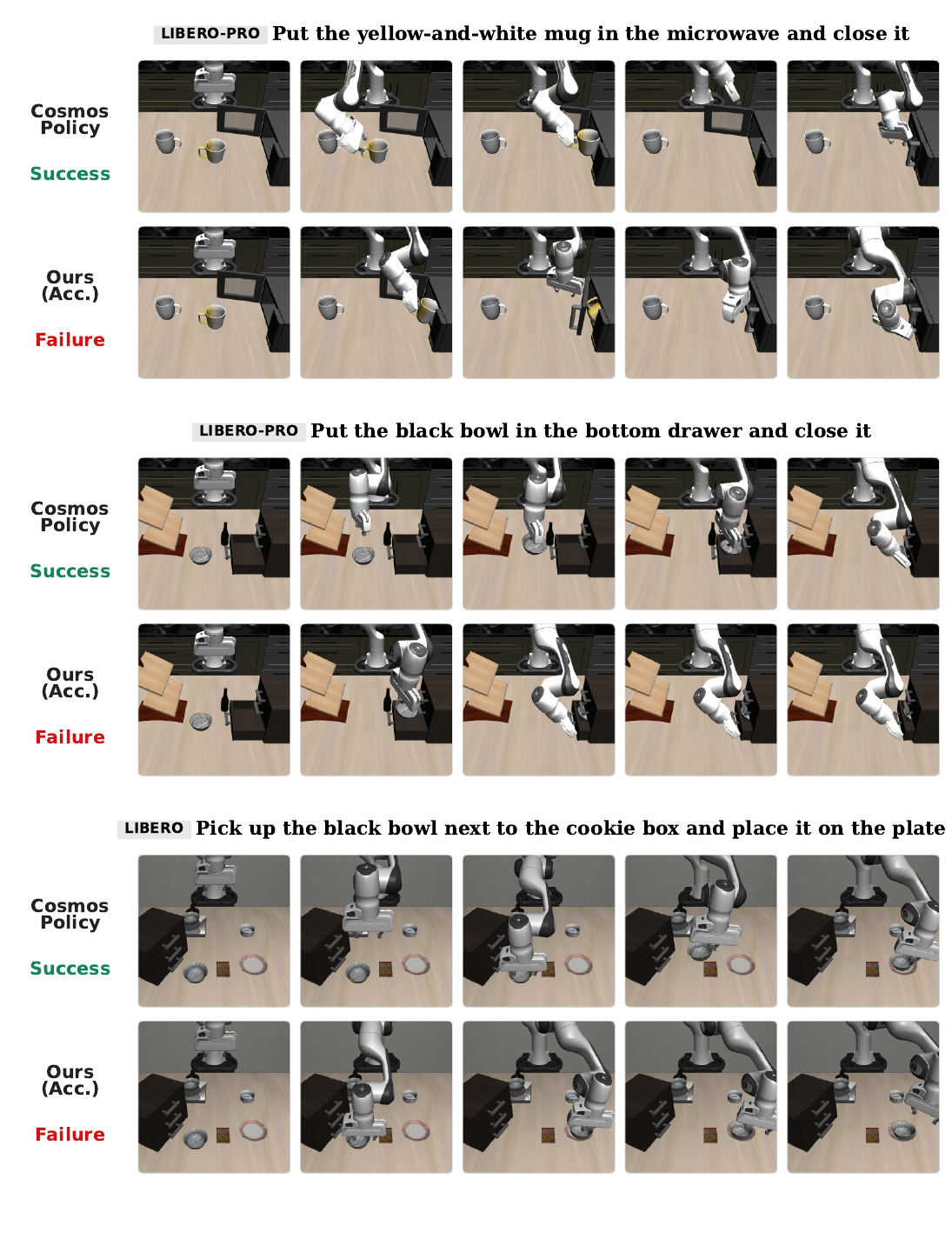}
    \vspace{-45pt}
    \caption{\textbf{Qualitative examples of failure cases under acceleration.}
    Representative comparisons in LIBERO and LIBERO-PRO where the original Cosmos Policy succeeds while Ours (Acc.) fails.}
    \label{fig:visual2}
    \vspace{-10pt}
\end{figure}

In contrast, Ours is relatively more sensitive to long-horizon sequential tasks involving drawers or microwaves, as illustrated in Figure~\ref{fig:visual2}. These tasks often require more precise grasping, positioning, and interaction, for which aggressive inference acceleration may introduce a slight degradation in control precision. Overall, these results suggest that our method largely preserves the original policy performance under substantial acceleration, with slightly more trajectories benefiting from acceleration than being degraded by it.

\vspace{-5pt}
\section{Efficiency--Performance Trade-off}
\label{sec:tradeoff}

\begin{figure}[t]
    \centering
    \includegraphics[width=\textwidth]{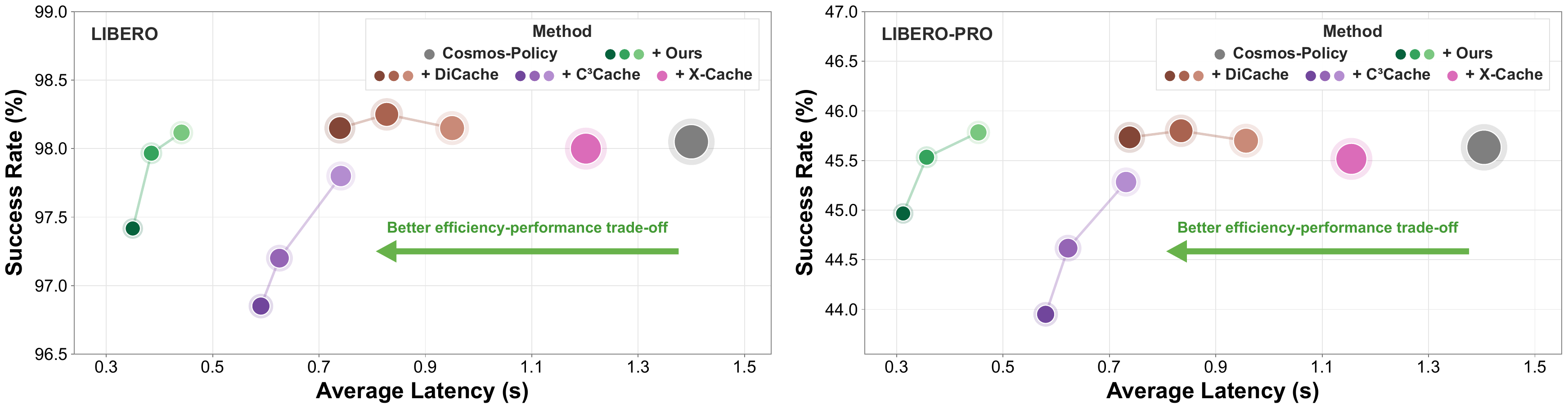}
    \caption{\textbf{Efficiency--performance trade-offs on LIBERO and LIBERO-PRO.} MARS achieves lower latency at comparable success rates.}
    \label{fig:liberoliberopro}
\end{figure}

We vary the caching strength of DiCache and C$^3$ache and evaluate multiple acceleration settings of MARS. Figure~\ref{fig:liberoliberopro} shows that MARS achieves lower latency at comparable success rates on both LIBERO and LIBERO-PRO. These results show that MARS offers a better efficiency--performance trade-off, with multiple operating points to meet different speed and task performance requirements.

\begin{table*}[t]
    \centering
    \caption{\textbf{Comparison of routing signals on LIBERO-PRO.}}
    \label{tab:routing_signals}
    \setlength{\tabcolsep}{7pt}
    \renewcommand{\arraystretch}{1.05}
    \resizebox{\textwidth}{!}{
    \begin{tabular}{lccccccc}
        \toprule
        \multicolumn{1}{l}{\textbf{LIBERO-PRO}}
        & \multirow{2}{*}{\textbf{\shortstack{Translation\\Threshold}}}
        & \multirow{2}{*}{\textbf{\shortstack{Rotation\\Threshold}}}
        & \multicolumn{4}{c}{\textbf{Success Rate (\%, $\uparrow$)}}
        & \multirow{2}{*}{\textbf{\shortstack{Average\\Success Rate} ($\uparrow$)}} \\
        \cmidrule(lr){4-7}
        \textbf{Routing Signal} & & & \textbf{Goal} & \textbf{Object} & \textbf{Spatial} & \textbf{Long} & \\
        \midrule
        Cosmos Policy & — & — & 43.67 & 46.13 & 47.40 & 45.33 & 45.63\% \\
        \midrule
        \rowcolor{rowmint}
        \textbf{Translation only} & 0.4 & — & 44.20 & 45.67 & 47.33 & 44.93 & \textbf{45.53\%} \\
        Translation + rotation & 0.4 & 0.02 & 44.13 & 45.47 & 47.07 & 44.00 & 45.17\% \\
        Translation + rotation & 0.4 & 0.05 & 44.00 & 45.53 & 47.00 & 43.80 & 45.08\% \\
        Translation + rotation & 0.4 & 0.10 & 43.93 & 45.67 & 47.33 & 44.73 & 45.42\% \\
        \bottomrule
    \end{tabular}
    }
\end{table*}

\section{Evaluation of Additional Routing Signals}
\label{sec:routing_signals}

We evaluate whether adding rotation magnitude improves translation-based routing. We fix the translation threshold at 0.4 and test rotation thresholds of 0.02, 0.05, and 0.10, selected around the median rotation magnitude. Table~\ref{tab:routing_signals} reports results on LIBERO-PRO averaged over three random seeds. Translation alone achieves the highest average success rate among the evaluated acceleration configurations, while adding rotation provides no further improvement. These results support translation magnitude as a simple and effective routing signal in the evaluated setting.

\section{Evaluation with Shared Configurations}
\label{sec:shared_configurations}

To evaluate configuration generality, we conduct additional experiments using shared configurations. All experiments in this section are performed on RTX 4090 GPUs, with results averaged over three random seeds. Across benchmarks, we use one fixed configuration on LIBERO, LIBERO-PRO, and RoboCasa without benchmark-specific tuning. Across backbones, we evaluate a shared configuration on Fast-WAM, LingBot-VA, and $\pi_{0.5}$ without backbone-specific tuning. Tables~\ref{tab:shared_benchmarks} and~\ref{tab:shared_backbones} report the corresponding results. Shared configurations consistently reduce latency while maintaining comparable success rates across the evaluated benchmarks and backbones.

\begin{table}[H]
    \centering
    \setlength{\tabcolsep}{7pt}
    \renewcommand{\arraystretch}{1.05}

    \caption{\textbf{Evaluation with a shared configuration across benchmarks.}}
    \label{tab:shared_benchmarks}
    \resizebox{0.8\linewidth}{!}{
    \begin{tabular}{lccccc}
        \toprule
        \multirow{2}{*}{\textbf{Benchmark}}
        & \multicolumn{2}{c}{\textbf{Success Rate (\%, $\uparrow$)}}
        & \multicolumn{2}{c}{\textbf{Latency (s, $\downarrow$)}}
        & \multirow{2}{*}{\textbf{Speedup ($\uparrow$)}} \\
        \cmidrule(lr){2-3}
        \cmidrule(lr){4-5}
        & \textbf{Base Model} & \textbf{Ours}
        & \textbf{Base Model} & \textbf{Ours} & \\
        \midrule
        LIBERO     & 98.05 & 98.37 & 0.8031 & \textbf{0.2491} & \textbf{3.22$\times$} \\
        LIBERO-PRO & 45.63 & 45.57 & 0.7800 & \textbf{0.2576} & \textbf{3.03$\times$} \\
        RoboCasa   & 66.69 & 66.22 & 1.3047 & \textbf{0.6137} & \textbf{2.13$\times$} \\
        \bottomrule
    \end{tabular}
    }

    \vspace{6pt}

    \caption{\textbf{Evaluation with a shared configuration across backbones.}}
    \label{tab:shared_backbones}
    \resizebox{0.8\linewidth}{!}{
    \begin{tabular}{lccccc}
        \toprule
        \multirow{2}{*}{\textbf{Backbone}}
        & \multicolumn{2}{c}{\textbf{Success Rate (\%, $\uparrow$)}}
        & \multicolumn{2}{c}{\textbf{Latency (s, $\downarrow$)}}
        & \multirow{2}{*}{\textbf{Speedup ($\uparrow$)}} \\
        \cmidrule(lr){2-3}
        \cmidrule(lr){4-5}
        & \textbf{Base Model} & \textbf{Ours}
        & \textbf{Base Model} & \textbf{Ours} & \\
        \midrule
        Fast-WAM    & 97.50 & 97.50 & 0.5295 & \textbf{0.1345} & \textbf{3.94$\times$} \\
        LingBot-VA  & 95.33 & 96.67 & 7.6555 & \textbf{1.4768} & \textbf{5.18$\times$} \\
        $\pi_{0.5}$ & 96.00 & 96.50 & 0.0829 & \textbf{0.0671} & \textbf{1.24$\times$} \\
        \bottomrule
    \end{tabular}
    }
\end{table}